\documentclass{article}
\PassOptionsToPackage{table}{xcolor}
\PassOptionsToPackage{sort&compress}{natbib}
\usepackage{iclr2027_conference,times}
\setcitestyle{numbers,square,citesep={,}}%
\usepackage{hyperref}
\usepackage{url}
\usepackage{graphicx}
\usepackage{float}
\usepackage{amssymb}
\usepackage{amsmath}
\usepackage{enumitem}
\usepackage{xcolor}
\usepackage{etoolbox}
\usepackage{caption}
\iclrfinalcopy

\usepackage{booktabs}
\usepackage{tabularx}
\usepackage{multirow}
\usepackage{xspace}
\usepackage{placeins}
\hypersetup{colorlinks=true,linkcolor=blue!45!black,citecolor=blue!45!black,urlcolor=blue!45!black}
\definecolor{astrablue}{HTML}{1764A0}
\definecolor{astralight}{HTML}{EAF2F9}
\colorlet{tabband}{black!8}
\newcommand{\best}[1]{\textbf{#1}}
\newcommand{\second}[1]{\underline{#1}}
\newcommand{\hd}[1]{\textbf{\boldmath #1}}
\AtBeginEnvironment{table}{}
\newcommand{\astra}{GPT-6-Astra\xspace}
\newcommand{\method}{SimplEvo\xspace}
\newcommand{\Mtask}{M^{\mathrm{task}}}
\newcommand{\Mwork}{M^{\mathrm{work}}}

\title{\fontsize{16.5}{20}\selectfont Review the Episodes, Refine the Lessons:\\Simple Memory Evolution for Embodied Agents}

\author{Yifei Yao$^{1*}$\quad Yude Zou$^{1,2*}$\quad Xing Gao$^{1\dagger}$\quad Guanbo Jiao$^{1,3}$\quad Yukai Wang$^{1}$\\
\bfseries Jianming Ma$^{1,2}$\quad Weinan Zhang$^{1,2}$\quad Chunhua Shen$^{1,4}$\\[3pt]
{\normalfont\small $^{1}$Shanghai Artificial Intelligence Laboratory\qquad $^{2}$Shanghai Jiao Tong University}\\
{\normalfont\small $^{3}$Fudan University\qquad $^{4}$Zhejiang University}\\[2pt]
{\normalfont\small $^{*}$Equal contribution\qquad $^{\dagger}$Corresponding author}}

\begin{document}
\fancypagestyle{firstpage}{%
  \fancyhf{}%
  \setlength{\headheight}{16pt}%
  \renewcommand{\headrulewidth}{0.4pt}%
  \lhead{\raisebox{-2pt}{\includegraphics[height=13pt]{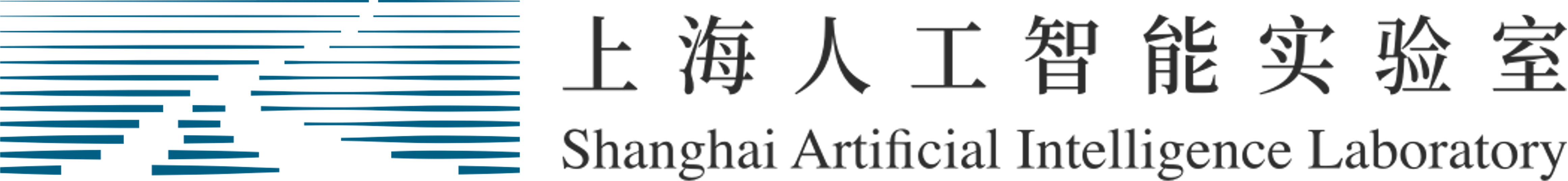}}}%
  \rhead{\raisebox{-2pt}{\includegraphics[height=12pt]{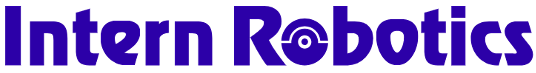}}}%
}
\maketitle
\thispagestyle{firstpage}
\lhead{Preprint}
\vspace{-10pt}
\begin{figure}[H]
\centering
\includegraphics[width=\linewidth]{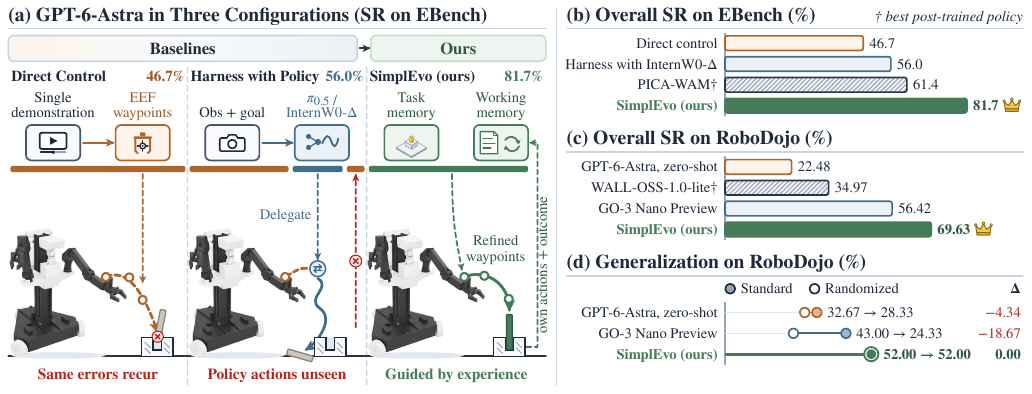}
\caption{\astra in two baseline configurations, direct control and a harness with policy, and with \method, whose working memory is written on training episodes and then fixed. (a) Each configuration with its EBench success rate (SR) and InternW0-$\Delta$ as the harness policy. The bar under each shows which component commands the robot, and the MLLM does not see the actions of a delegated policy. (b) Overall SR on EBench and (c) on RoboDojo, where $\dagger$ marks the strongest post-trained policy and zero-shot denotes the official entry of \astra. (d) SR on the 12 RoboDojo Generalization tasks with standard and randomized configurations.}
\label{fig:teaser}
\end{figure}

\begin{abstract}
Recent studies of embodied agents based on frontier multimodal large language models (MLLMs) develop harness engineering around the MLLM, such as harnesses with post-trained policies and recursive self-improvement loops. Yet whether these additions exceed the native capability of the MLLM remains unexplored. We introduce \method to explore how far an MLLM can improve under direct control when it retains its execution experience in self-written text memories. It requires no weight updates, post-trained policies, or privileged information. The MLLM writes a coordinate-free \textit{task memory} once from a single demonstration. Under a strict train--test protocol, it then creates a bounded \textit{working memory} on training episodes only and refines it after each episode with lessons on procedure, completion, and recovery. Both memories stay fixed at test time. On EBench, \method lifts the success rate (SR) of \astra under direct control from 46.7\% to 81.7\%, and tops the RoboDojo leaderboard with 69.63\% SR and a score of 75.32. Compared with memories that also store past visual frames, \method reaches a similar SR with text memory alone while using fewer tokens. With its memories fixed, \method also generalizes to unseen layouts and randomized scene configurations. Moreover, the working memory transfers across models: written by \astra, it can be reused as is by weaker MLLMs and improve their performance without any adaptation of their own.
\end{abstract}

\section{Introduction}
\label{sec:intro}

Generalist robotic manipulation is a central problem of embodied intelligence. Vision-language-action models (VLAs)~\citep{kim2024openvla, black2025pi0, pi05} and world action models (WAMs)~\citep{dreamzero, yuan2026fastwam, wang2026openwam} align their manipulation skills through post-training on benchmark-specific trajectories~\citep{liu2023libero, mu2025robotwin, chen2025robotwin}. In contrast, recent studies show that frontier multimodal large language models (MLLMs) directly operate robots to complete tasks without parameter updates. They command the end effector from camera views and language instructions~\citep{openai2026astra, zhang2026robodojo, liu2026roboicl, jia2026agent, lou2026knowbody, hu2026via}, explore and adapt their behaviors to the observed results~\citep{guo2026robodawn, cheng2026gptpolicy}, and recover from errors within an episode~\citep{fu2026capx, yao2026profiling}. These findings have motivated a growing range of embodied agents built on frontier MLLMs.

These agents fall into two lines of work. The first is \emph{code-as-policies}, which began with language models writing robot code~\citep{codeaspolicies} and composes perception modules and motion primitives into executable programs. In systems such as Code as Policies~\citep{codeaspolicies}, COAP~\citep{hu2026embodied}, and RAX on RoboDojo~\citep{huang2026rax}, the synthesized programs run at test time without the MLLM in the loop, so they cover only anticipated scene variations and failure modes. Recent works such as CaP-X~\citep{fu2026capx} and PyRUA-Lean~\citep{si2026pyrua} therefore execute code online with the MLLM in the loop. Even so, code-as-policies systems typically rely on privileged information beyond RGB views and proprioception, such as metric depth, camera parameters, and object sizes, or on estimates from calibrated depth sensors and vision foundation models~\citep{kirillov2023sam, ravi2025sam2, yang2024depthanything, yang2024depthanythingv2}, to which the resulting programs are sensitive. Frontier MLLMs, by contrast, already perceive and reason about space from RGB views and proprioception, which raises the question of whether routing perception through intermediate models is necessary at all.

The second is \emph{harness engineering}, which extends the harness around the MLLM with post-trained policies or external improvement loops. Some systems build \emph{harnesses with policy} that expose post-trained policies as callable tools~\citep{zhang2026harnessvla, wang2026harnesspai}, so the MLLM plans while the policy executes low-level actions. Each delegated call returns only its final observation and status, so the actions of the policy never enter the context of the MLLM, and the harness can constrain the capability of the underlying MLLM. Other systems adapt self-evolving language agents into embodied \emph{recursive self-improvement} (RSI) frameworks that revise context, tools, or control logic through revision loops around the agent~\citep{tian2026physevo, physicalrsi2026, lu2026aspire, cheng2026asena}. These frameworks iterate on external harnesses and rarely measure the base model under direct control first. Some self-evolving designs adapt on the test scenes themselves~\citep{wang2026rethinking}. Because the evaluator then returns feedback on the scenes used for evaluation, test-scene information can leak into the evolved system, and its gains no longer show whether the acquired experience generalizes to scenes the agent has never attempted. It thus remains unclear how far a frontier MLLM can go under direct control on its own, and hence how much of this added machinery is necessary.

Both lines of work focus on either exploiting the coding ability of the MLLM or wrapping it with post-trained policies and improvement loops, thereby overlooking the rising capability of the MLLM itself to act directly as a general-purpose robot controller. We refer to this setting as \emph{direct control}, where the MLLM commands the end effector directly using RGB views and proprioception, without external perception modules or a post-trained policy. Direct control returns the result of every executed command to the MLLM, so the closed-loop feedback and within-episode recovery are explicit. After a missed grasp or a failed insertion, the MLLM compares the commanded end-effector coordinates with the resulting camera views, diagnoses the error, and adjusts its next command~\citep{yao2026profiling}. These adjustments and the experience of each trial, however, vanish when the episode ends, so each episode starts anew. Can a frontier MLLM unlock strong manipulation under direct control simply by inheriting and evolving the in-episode adjustments and trial experience across episodes, relying only on its native multimodal capabilities?

To show that recording and revising its own memory is enough for a frontier MLLM to excel under direct control, we present \method, illustrated in Figure~\ref{fig:framework}. \method keeps the native direct-control loop unchanged and introduces only two compact text memories that the MLLM writes itself during adaptation on the training set. From one training demonstration, the MLLM formulates a coordinate-free \emph{task memory} once, and after each training episode, it reflects on its own trajectory in one turn to rewrite a bounded \emph{working memory} of transferable lessons. A strict train--test protocol confines adaptation and its outcome feedback to training scenes, none of which appears in the test set. Both memories are then fixed, and every test episode reads them without reflection turns or memory updates. Throughout adaptation and testing, the agent observes only RGB views and proprioception and requires no post-trained policies, weight updates, or privileged information.

With its working memory fixed after adaptation, \method outperforms post-trained policies and other MLLM agents on the test episodes of EBench~\citep{gao2026ebench} and the test layouts of RoboDojo~\citep{chen2026robodojo}. On EBench, \method raises \astra from 46.7\% under direct control to a success rate (SR) of 81.7\% across its 510 test scenes while reducing tokens per success, against up to 56.0\% SR for harnesses with policy and 61.4\% SR for the strongest post-trained policy. On RoboDojo, after adapting on 15 training layouts per task, \method leads the public leaderboard across 50 test layouts per task with 69.63\% SR and a score of 75.32.

This paper makes three contributions:
\begin{itemize}[leftmargin=*,itemsep=1pt,topsep=2pt]
  \item We propose \method, a two-stage memory framework for direct control, in which the MLLM derives a task memory without coordinates from a single training demonstration and rewrites a bounded working memory through its own reflection on every training episode, without post-trained policies or weight updates.
  \item Under a strict evaluation protocol with train--test separation, native step limits, and no privileged information, \method ranks first on both the EBench and the RoboDojo leaderboards, ahead of post-trained VLAs, WAMs, and other MLLM agents, while using fewer tokens per success than direct control.
  \item We show that \method performs well with simple text-based memories and, with these memories fixed, generalizes to unseen layouts and randomized scene configurations. Moreover, the working memory created by a frontier MLLM can be reused by weaker MLLMs to improve their performance.
\end{itemize}

\section{Related Work}
\label{sec:related}

\paragraph{Harnesses and self-improvement for embodied agents.} Beyond the code-as-policies systems and harnesses with policy discussed in Section~\ref{sec:intro}, harnesses also coordinate post-trained policies through memory~\citep{baton}, and direct control can command end-effector targets~\citep{zhang2026robodojo, jia2026agent, lou2026knowbody} or semantic action units~\citep{chen2026showharness}. Self-improving embodied systems differ in what they change. They revise tools, harnesses, or code policies during development and score the fixed result~\citep{tian2026physevo, physicalrsi2026, hu2026embodied}, co-evolve guidance with harness modules~\citep{wang2026shaper, ding2026zetta, li2026racap, tang2026adahvla, wang2026regenharness}, or distill trajectories into policy weights~\citep{robocat, recap, xiao2026enpire}, and \citet{zhou2026whoimproves} classify such systems by which component a retained change updates and how broadly its gain holds. Controlled audits find that harness evolution does not consistently outperform test-time scaling at equal budget, that frequent updates can fail to improve held-out performance~\citep{wang2026rethinking, zheng2026seagym}, and that selection on small, noisy evaluations admits spurious gains~\citep{shawn2026pace, gideoni2026simple}. \method keeps tools, harnesses, code, and model weights fixed. It acts under direct control and updates only the text working memory that the MLLM reads across episodes.

\paragraph{Frontier MLLMs as robot policies.} Frontier MLLMs now control robots without being trained on robot data. For example, \astra~\citep{openai2026astra} commands end effectors directly on RoboDojo~\citep{chen2026robodojo, zhang2026robodojo} and learns from demonstration blocks in context~\citep{liu2026roboicl}, while other MLLM agents write robot commands from saved procedures~\citep{jia2026agent} or refine a revisable model of the robot body~\citep{lou2026knowbody}. \citet{yao2026profiling} profile such an agent on EBench~\citep{gao2026ebench} across direct control and harnesses with policy, and they find that its failures concentrate in physical execution. This study measures how far an agent under direct control improves once its execution experience persists across episodes.

\paragraph{Memory and experience evolution.} Agent memory systems differ in how they represent and update experience across episodes. Language agents improve without weight updates by accumulating verbal reflections~\citep{shinn2023reflexion}, extracting insights from successful and failed trajectories~\citep{zhao2024expel}, or curating evolving playbooks~\citep{zhang2026agentic}. In multimodal interactive environments, VISTA~\citep{han2026vista} preserves a lossless visual history of past observations, so carrying it across manipulation episodes lengthens the context with every executed command. Embodied agents store experience as executable skills~\citep{voyager}, validated program repairs~\citep{lu2026aspire}, policy repositories of prompts, tools, and control code~\citep{elmaaroufi2026rho}, per-task programs~\citep{wang2026harnesspai}, or a knowledge package that \astra writes once from training demonstrations and reads unchanged in every test episode~\citep{sensenova2026astra}. Unlike reflections, insights, and playbooks distilled from language-level actions, or knowledge written from demonstrations, the working memory of \method distills the low-level commands that the MLLM itself issued, is rewritten in full within a fixed size after each adaptation episode, and stays fixed at test time.

\section{Method}
\label{sec:method}

\begin{figure*}[t]
  \centering
  \includegraphics[width=\linewidth]{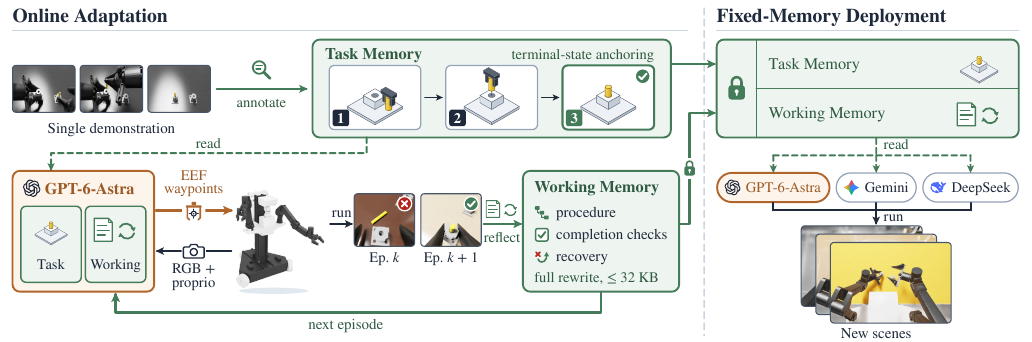}
  \caption{Overview of \method. During online adaptation (left), an annotator instance of \astra reviews one training demonstration and writes a coordinate-free task memory of the stages and the completed state, and after every training episode, \astra rewrites the working memory from the trajectory and outcome of that episode into lessons on procedure, completion, and recovery. Once fixed (right), both memories are read in new scenes, and other MLLMs read the working memory of \astra without modification. All panels show peg insertion on EBench.}
  \label{fig:framework}
\end{figure*}

As shown in Figure~\ref{fig:framework}, \method equips direct control with two memories written by the MLLM, each overcoming a limitation of isolated episodes. The \emph{task memory} supplies the order of stages and the final state of a task before the first attempt, and the \emph{working memory} preserves the corrections that would otherwise vanish when an episode ends.

\subsection{Direct Control under a Strict Information Protocol}
\label{sec:method:setting}
\paragraph{Agent under direct control.} The agent pairs a fixed-weight MLLM core $F$ with an agent harness that specifies its observation and action space. Under \emph{direct control}, the MLLM issues every robot command itself instead of calling a post-trained policy. At decision step $t$ of the $k$-th episode on task $\tau$, the MLLM receives the observation $o_t=(I_t, q_t)$, which consists of RGB images $I_t$ from an overview camera and two wrist cameras, or from any subset of them that the MLLM requests, and the proprioceptive state $q_t$ of the robot. Given the observations $o_{\le t}$ and the earlier actions $a_{<t}$ of the episode, the MLLM issues
\begin{equation}
  a_t = F\big(\ell_\tau,\, \Mtask,\, \Mwork_{k-1},\, o_{\le t},\, a_{<t}\big),
  \label{eq:policy}
\end{equation}
where $\ell_\tau$ is the task instruction with basic specifications of the robot embodiment, and $\Mtask$ and $\Mwork_{k-1}$ are the task and working memories of Sections~\ref{sec:method:task} and~\ref{sec:method:working}. Each action $a_t$ is a chunk of 1 to 6 absolute waypoints, each setting an end-effector pose and a gripper target for each arm and, on the mobile robot of EBench, a base pose. Inverse kinematics converts the end-effector poses into joint commands. An episode ends when the evaluator registers success, when the step limit of the task is reached, or earlier when the MLLM ends its turn. The observations and actions of episode $k$ form its trajectory $\xi_k=(o_1,a_1,o_2,a_2,\dots)$. By comparison, a \emph{harness with policy} hands control to a post-trained policy for a bounded number of physics steps and receives only the end state, so $\xi_k$ omits the actions that the policy executes.

\paragraph{Information protocol.} On both benchmarks, we restrict the agent to the information that a physical deployment provides. During execution, the agent observes the scene solely through RGB images and robot proprioception $o_t=(I_t, q_t)$, matching the inputs of a post-trained VLA or WAM. Although recent evaluation guidelines for robot-use agents allow camera intrinsics~\citep{openrua2026revisiting}, we exclude privileged information such as depth, camera parameters, or object states used by prior systems~\citep{codeaspolicies, hu2026embodied, huang2026rax}. On a physical robot, such information would come from external estimation modules that are sensitive to perception errors. Restricting observations to RGB views and proprioception therefore reflects more faithfully how the agent adapts to the robot embodiment and refines its execution in the environment. Beyond $o_t$, the MLLM receives only the instruction $\ell_\tau$, the two memories, and the training demonstration $D_\tau$ of Section~\ref{sec:method:task}. Finally, like post-trained policies, the agent acts within the native per-task step limits of each benchmark and receives no intermediate progress signals.

\subsection{Task Memory from a Demonstration}
\label{sec:method:task}
The \emph{task memory} $\Mtask$ records the stage order and the visual completion criterion that a single demonstration conveys reliably. Before the first episode, an annotator instance of the same MLLM reviews the recorded actions and camera frames of the demonstration $D_\tau$ and writes $\Mtask$ once. $\Mtask$ lists the stages of the task in order, stating for each stage the operation and the arm that performs it, and ends with the visual state of the completed task. It contains no coordinates, so that it applies to layouts other than the demonstrated one. Because the final stage specifies the completed state, $\Mtask$ gives the MLLM an explicit stopping criterion before any execution experience exists, which we term \emph{terminal-state anchoring}. The MLLM reads $\Mtask$ in every episode and can revisit $D_\tau$ during execution. Tasks without a demonstration, the eight Open tasks of RoboDojo, run without $\Mtask$. Appendix~\ref{sec:supp:protocol:annotation} provides the annotation prompt and the tools for reviewing the demonstration.

\subsection{Working Memory via Trajectory Reflection}
\label{sec:method:working}
The \emph{working memory} $\Mwork$ carries execution experience across the episodes of a task through compact text rewrites. After episode $k$, the same MLLM $F$ updates the working memory in a single reflection turn without robot actions:
\begin{equation}
  \Mwork_{k} = F\big(\Mwork_{k-1},\, \xi_k,\, r_k\big), \qquad |\Mwork_{k}| \le B,
  \label{eq:update}
\end{equation}
where $\xi_k$ is the executed trajectory and $r_k$ is the terminal outcome, the success and score reported by the benchmark evaluator as an episode-level reward. In each update, the MLLM may replace $\Mwork_{k-1}$ with a full rewrite within a size budget $B$ of 32\,KB, so the MLLM itself decides what the bounded memory keeps. Replacing the memory in full allows the MLLM to revise or discard outdated content while keeping $\Mwork_{k}$ far shorter than the cumulative history of the chain. By analyzing the executed commands and their observed consequences in $\xi_k$, particularly the causes of failures, the MLLM distills concise lessons on task procedure, completion criteria, and error recovery, accompanied by representative commands and outcomes from earlier attempts as supporting evidence. To discourage path dependence on the MLLM's own exploration trajectories, the description of the memory-writing tool asks it to treat measured coordinates as scene-specific evidence instead of commands to replay in another scene. Appendix~\ref{sec:supp:reflection} gives the reflection prompt and the memory tools, and Appendix~\ref{sec:supp:workmem} presents excerpts of working memories.

\subsection{Online Adaptation and Deployment}
\label{sec:method:deploy}
During online adaptation, the agent executes the episodes of a task sequentially along an adaptation \emph{chain}, starting with the working memory $\Mwork_{0}$ empty. Under the \emph{train--test protocol}, online adaptation runs exclusively on training scenes or layouts, after which the working memory $\Mwork_K$ at the end of the chain is fixed. At deployment, $\Mwork_K$ takes the place of $\Mwork_{k-1}$ in \eqref{eq:policy} in every new scene without further updates, so no information from deployment scenes can leak into the memory. Appendix~\ref{sec:supp:chains} analyzes the dynamics of the adaptation chains.

\section{Experiments}
\label{sec:experiments}
We evaluate \method on two public manipulation benchmarks against their leaderboard entries and against \astra without working memory, with adaptation confined to training scenes.

\subsection{Setup}
\label{sec:setup}
\paragraph{Benchmarks and generalization.} The test episodes of EBench~\citep{gao2026ebench} and RoboDojo~\citep{chen2026robodojo} evaluate the working memory under variations in scene conditions. EBench comprises 26 tasks of a simulated ARX Lift~2 dual-arm mobile robot in Isaac Sim~\citep{gao2025genmanip}, and its 510 test scenes perturb the manipulated objects, the background, the instruction, or several of them at once. Its seven single-layout tasks keep one object layout across training and test episodes while varying objects, backgrounds, and instructions. Its 19 multi-layout tasks also vary the layout, so a test layout may be unseen or already seen in training. Fifteen of these 19 tasks include both seen and unseen test layouts. RoboDojo comprises 42 dual-arm tabletop tasks in five capability dimensions, each tested on 50 layouts of initial objects and their poses. On its 12 Generalization tasks, half of the test layouts randomize the background, the lighting, the target objects, and the clutter. Following the official protocols, we report SR and the stage-completion score, both on a 0--100 scale.

\paragraph{Train--test protocol.} Per task, adaptation uses up to 15 EBench training episodes or 15 RoboDojo training episodes, then fixes the working memory for testing. On EBench, adaptation stops early after three consecutive successes, as Appendix~\ref{sec:supp:chains} details. On RoboDojo Generalization tasks, training includes both standard and randomized configurations. The task memory $\Mtask$ comes from one training demonstration per task.

\paragraph{Agent and baselines.} We use \astra~\citep{openai2026astra} at high reasoning effort as the MLLM core $F$ in all experiments except the cross-model transfer study. On EBench, we compare against the 12 post-trained VLAs and WAMs on the public leaderboard, as well as \astra under direct control and in harnesses with $\pi_{0.5}$~\citep{pi05} or InternW0-$\Delta$~\citep{miao2026internw0deltaworldactionmodel} following \citet{yao2026profiling}, all reading the task memory but no working memory. On RoboDojo, we compare against the leading post-trained policies and agents on the public leaderboard, the official zero-shot entry of \astra on test layouts, and task memory only on the training layouts used by \method. We also report tokens per success, the total tokens of an evaluation divided by its successes, and analyze cost in Appendix~\ref{sec:supp:cost}.

\subsection{Results on EBench}
\label{sec:results:ebench}
With its working memory fixed, \method reaches 81.7\% SR on the 510 test episodes of EBench and ranks first in every task group of Table~\ref{tab:ebench}, using the leaderboard snapshot of 29 October 2026, ahead of PICA-WAM~\citep{zou2026picawam}, the best post-trained policy, at 61.4\% and the best harness with policy at 56.0\%. The largest gains arise on high-precision, tabletop, and long-horizon tasks. On high-precision tasks, direct control without working memory succeeds in 11.3\% of the episodes, and neither harness with policy exceeds 12.5\%. With working memory, the same direct control reaches 82.5\%, 2.3 times the 36.3\% of PICA-WAM, which shows that the waypoints of the MLLM are precise enough for these tasks once the corrections of earlier attempts persist. On long-horizon tasks, direct control drops to 27.3\% from 73.2\% on short-horizon ones, while \method reaches 76.3\%, above even the short-horizon SR of direct control. The task memory fixes the stage sequence, and the working memory adds to each stage the completion checks and recovery steps verified in earlier episodes. On bottle collection, \method reaches 45.0\% SR, while none of the 12 leaderboard policies succeeds. It also reaches 90.0\% on coffee-bean collection and 95.0\% on shop checkout, where direct control without working memory has no successes. At test time, \method spends 2.76M tokens per success, compared with 8.12M for direct control, 3.46M for the harness with $\pi_{0.5}$, and 3.99M for the harness with InternW0-$\Delta$, as Appendix~\ref{sec:supp:cost} reports.

\begin{table*}[t]
  \centering
  \footnotesize
  \setlength{\tabcolsep}{1.5pt}
  \caption{SR (\%) and score on the 510 test episodes of EBench for the post-trained policies on the public leaderboard (29 October 2026) and for \astra without working memory and with \method. Bold and underline mark the best and second-best values per column.}
  \label{tab:ebench}
  \begin{tabular*}{\linewidth}{@{\extracolsep{\fill}}l*{9}{c}@{}}
\toprule
 & \multicolumn{2}{c}{\hd{Overall}} & \multicolumn{3}{c}{\hd{Precision}} & \multicolumn{2}{c}{\hd{Mobility}} & \multicolumn{2}{c}{\hd{Horizon}} \\
\cmidrule(lr){2-3}\cmidrule(lr){4-6}\cmidrule(lr){7-8}\cmidrule(lr){9-10}
\hd{System} & \hd{SR} & \hd{Score} & \hd{Low} & \hd{Med.} & \hd{High} & \hd{Mobile} & \hd{Tabletop} & \hd{Short} & \hd{Long} \\
\midrule
\rowcolor{tabband} \multicolumn{10}{c}{\textit{Post-trained policies}} \\
\midrule
X-VLA~\citep{zheng2025xvla} & 23.7 & 35.8 & 35.7 & 13.3 & 2.5 & 29.3 & 8.6 & 45.9 & 7.4 \\
Fast-WAM~\citep{yuan2026fastwam} & 25.6 & 37.1 & 35.7 & 19.0 & 3.8 & 34.0 & 2.9 & 46.4 & 10.4 \\
Cosmos3-Edge~\citep{nvidia2026cosmos3} & 29.3 & 42.5 & 38.6 & 20.8 & 13.8 & 33.2 & 18.6 & 39.1 & 22.1 \\
GigaBrain-0.7~\citep{gigabrainteam2026gigabrain} & 33.3 & 46.0 & 35.4 & 33.8 & 25.0 & 31.6 & 37.9 & 40.0 & 28.3 \\
$\pi_0$~\citep{black2025pi0} & 33.7 & 47.5 & 39.0 & 30.0 & 22.5 & 34.0 & 32.9 & 40.5 & 28.8 \\
InternVLA-A1.5~\citep{ma2026internvla} & 34.2 & 46.8 & 44.6 & 26.0 & 13.8 & 42.3 & 12.1 & 46.8 & 24.9 \\
$\pi_{0.5}$~\citep{pi05} & 41.4 & 54.4 & 49.6 & 39.0 & 17.5 & 45.9 & 29.3 & 52.3 & 33.4 \\
Qwen-RobotManip~\citep{yuan2026qwenrobotmanip} & 45.6 & 60.8 & 50.6 & 50.2 & 18.8 & 43.9 & \second{50.0} & 55.9 & 38.0 \\
InternW0-$\Delta$~\citep{miao2026internw0deltaworldactionmodel} & 49.2 & 66.0 & 58.7 & 49.2 & 16.3 & 55.0 & 33.6 & 59.1 & 42.0 \\
Vela~\citep{li2026vela} & 49.7 & 65.5 & 59.5 & 45.4 & 23.8 & 51.7 & 44.3 & 59.5 & 42.4 \\
OpenWAM-$\alpha$~\citep{wang2026openwam} & 55.3 & 70.1 & 67.4 & 45.6 & 32.5 & 60.2 & 42.1 & 65.9 & 47.6 \\
PICA-WAM~\citep{zou2026picawam} & \second{61.4} & \second{74.9} & \second{77.5} & 45.8 & \second{36.3} & \second{67.5} & 45.0 & 68.6 & \second{56.1} \\
\midrule
\rowcolor{tabband} \multicolumn{10}{c}{\textit{Agents based on \astra}} \\
\midrule
Direct control~\citep{yao2026profiling} & 46.7 & 65.4 & 60.6 & 40.2 & 11.3 & 56.6 & 20.0 & 73.2 & 27.3 \\
Harness with $\pi_{0.5}$~\citep{yao2026profiling} & 53.0 & 69.2 & 68.9 & 47.9 & 7.5 & 61.0 & 31.4 & 70.5 & 40.2 \\
Harness with InternW0-$\Delta$~\citep{yao2026profiling} & 56.0 & 74.4 & 68.2 & \second{56.5} & 12.5 & 64.3 & 33.6 & \second{77.3} & 40.4 \\
\rowcolor{astralight} \method (ours) & \best{81.7} & \best{91.2} & \best{90.1} & \best{66.7} & \best{82.5} & \best{83.2} & \best{77.9} & \best{89.1} & \best{76.3} \\
\bottomrule
\end{tabular*}

\end{table*}

\begin{table*}[t]
  \centering
  \footnotesize
  \setlength{\tabcolsep}{2pt}
  \caption{SR (\%) and score on the 42 tasks of RoboDojo, over 50 layouts per task, for entries on the public leaderboard (2 November 2026) and for \method on its test layouts. The w/o Open columns average the four dimensions other than Open, as on the leaderboard. Bold and underline mark the best and second-best values.}
  \label{tab:robodojo}
  \begin{tabular*}{\linewidth}{@{\extracolsep{\fill}}ll*{9}{c}@{}}
\toprule
 & & \multicolumn{2}{c}{\hd{Overall}} & \multicolumn{2}{c}{\hd{w/o Open}} & \multicolumn{5}{c}{\hd{Capability SR (\%)}} \\
\cmidrule(lr){3-4}\cmidrule(lr){5-6}\cmidrule(lr){7-11}
\hd{System} & \hd{Class} & \hd{SR} & \hd{Score} & \hd{SR} & \hd{Score} & \hd{Gen.} & \hd{Prec.} & \hd{Long} & \hd{Mem.} & \hd{Open} \\
\midrule
\rowcolor{tabband} \multicolumn{11}{c}{\textit{Post-trained policies}} \\
\midrule
$\pi_{0.5}$~\citep{pi05} & VLA & 6.93 & 11.44 & 8.25 & 13.80 & 8.17 & 5.50 & 14.67 & 4.67 & 1.67 \\
Awomo-0.5~\citep{awomowam2026awomo} & WAM & 29.64 & 35.34 & 31.33 & 38.15 & 23.84 & 33.50 & 26.75 & 41.22 & 22.92 \\
VPP2~\citep{guo2026vpp2} & WAM & 32.26 & 39.26 & 39.31 & 47.93 & 29.50 & 35.17 & 37.00 & 55.56 & 4.08 \\
WALL-OSS-1.0-lite~\citep{zhai2025igniting} & VLA & 34.97 & 42.59 & 38.82 & 48.05 & 37.94 & 29.08 & 32.25 & 56.00 & 19.58 \\
\midrule
\rowcolor{tabband} \multicolumn{11}{c}{\textit{Agents}} \\
\midrule
\astra, zero-shot~\citep{zhang2026robodojo} & Agent & 22.48 & 28.97 & 20.36 & 27.63 & 30.50 & 4.00 & 8.25 & 38.67 & 31.00 \\
WorldScape-R$^2$SI~\citep{manifold2026r2si} & Agent + WAM & 39.35 & 46.58 & 37.31 & 45.93 & 35.83 & 15.50 & 34.25 & 63.67 & 47.50 \\
HexaShell-Preview~\citep{gao2026hexashell} & Agent & 46.85 & 53.18 & 44.75 & 51.98 & 34.50 & 32.00 & 31.50 & \second{81.00} & 55.25 \\
PyRUA-Lean~\citep{si2026pyrua} & Agent & 50.35 & 55.26 & 48.37 & 53.93 & \second{43.17} & 38.25 & 43.75 & 68.33 & \second{58.25} \\
GO-3 Nano Preview & Agent + WAM & \second{56.42} & \second{61.90} & \second{56.71} & \second{61.95} & 33.66 & \best{76.00} & \second{53.50} & 63.67 & 55.25 \\
\rowcolor{astralight} \method (ours) & Agent & \best{69.63} & \best{75.32} & \best{67.67} & \best{74.09} & \best{52.00} & \second{64.50} & \best{58.50} & \best{95.67} & \best{77.50} \\
\bottomrule
\end{tabular*}

\end{table*}

\begin{table}[t]
  \centering
  \footnotesize
  \setlength{\tabcolsep}{3pt}
  \caption{SR (\%) and score of \astra on the 42 tasks of RoboDojo: task memory only and online adaptation on the 15 training layouts per task, and the fixed working memory of \method on the 50 test layouts per task. Bold marks the best value in each column.}
  \label{tab:robodojo_stages}
  \begin{tabularx}{\linewidth}{ll*{5}{>{\centering\arraybackslash}X}}
\toprule
 & & \multicolumn{2}{c}{\hd{Overall}} & \multicolumn{2}{c}{\hd{w/o Open}} & \hd{Open} \\
\cmidrule(lr){3-4}\cmidrule(lr){5-6}\cmidrule(lr){7-7}
\hd{Configuration} & \hd{Layouts (Episodes)} & \hd{SR} & \hd{Score} & \hd{SR} & \hd{Score} & \hd{SR} \\
\midrule
Task memory only & 15 training (630) & 36.37 & 43.69 & 37.34 & 45.78 & 32.50 \\
\method, online & 15 training (630) & 58.53 & 65.04 & 55.66 & 63.15 & 70.00 \\
\midrule
\rowcolor{astralight} \method, fixed working memory & 50 test (2{,}100) & \best{69.63} & \best{75.32} & \best{67.67} & \best{74.09} & \best{77.50} \\
\bottomrule
\end{tabularx}

\end{table}

\subsection{Results on RoboDojo}
\label{sec:results:robodojo}
After adapting on the 15 training layouts of each task, \method ranks first among the RoboDojo leaderboard entries of Table~\ref{tab:robodojo} with 69.63\% SR and a score of 75.32 on the 50 test layouts per task. It also has the highest SR and score when the Open dimension is excluded. Table~\ref{tab:robodojo_stages} shows two effects of the working memory. First, reflection distills experience into working memory on the eight Open tasks, which provide no demonstration or task memory. On the training layouts, SR rises from 32.50\% without working memory to 70.00\% during adaptation. After adaptation, the fixed working memory reaches 77.50\% on the test layouts. Second, the fixed working memory carries training experience into test episodes. Test SR exceeds the 58.53\% observed during online adaptation, and tokens per success are 2.23M on test layouts, compared with 4.12M for task memory alone on training layouts. Appendix~\ref{sec:supp:agentic} compares agentic systems outside the leaderboard.

\section{Analysis}
\label{sec:analysis}
\begin{table}[t]
  \centering
  \footnotesize
  \setlength{\tabcolsep}{3pt}
  \caption{Memory designs of \astra under direct control on one online chain of ten validation scenes per high-precision EBench task: SR (\%) per task and overall, mean score, tokens per episode (M), and minutes of control per episode at equal simulator speed (model time plus 0.5\,s per physics step). The visual-replay row includes text working memory, and VISTA history keeps the whole chain in one conversation. Bold marks the best value in each column.}
  \label{tab:design}
  \newcolumntype{W}[1]{>{\hsize=#1\hsize\centering\arraybackslash}X}%
\begin{tabularx}{\linewidth}{l W{1.05} W{0.85} W{0.85} W{0.85} W{0.85} W{1.05} W{1.15} W{1.35}}
\toprule
\hd{Memory} & \hd{Coffee} & \hd{Gear} & \hd{Peg} & \hd{Nut} & \hd{SR} & \hd{Score} & \hd{Tokens} & \hd{Minutes} \\
\midrule
Task memory only & 0.0 & 40.0 & 10.0 & 10.0 & 15.0 & 40.5 & 3.51 & 29.5 \\
\rowcolor{astralight} + text working memory (\method) & 20.0 & 70.0 & \best{90.0} & 10.0 & \best{47.5} & 64.1 & \best{2.86} & \best{23.0} \\
Text working memory + visual replay & \best{30.0} & 70.0 & 70.0 & 20.0 & \best{47.5} & 64.5 & 3.47 & 27.2 \\
VISTA history~\citep{han2026vista} & 20.0 & \best{90.0} & 40.0 & \best{30.0} & 45.0 & \best{68.0} & 6.10 & 31.4 \\
\bottomrule
\end{tabularx}

\end{table}

\subsection{Design Choices for Memory and Reflection}
\label{sec:analysis:format}
\paragraph{Text versus visual memory.} On online chains of ten validation scenes for each of the four high-precision EBench tasks, the task memory stays fixed and only the working-memory representation varies. Each chain starts from an empty working memory, so its SR includes the first attempts of adaptation, unlike the fixed working memory of Table~\ref{tab:ebench}. Under direct control, Table~\ref{tab:design} shows that the text working memory more than triples the SR of task memory alone, from 15.0\% to 47.5\%. Adding a visual replay of earlier attempts, which archives their camera frames, to the text working memory leaves the SR unchanged, and the MLLM queries the replay in only 3 of the 40 episodes. Maintaining a single continuous conversation across the chain, following the visual history design of VISTA~\citep{han2026vista}, reaches a similar SR with a slightly higher score. The text working memory uses the fewest tokens per episode and the shortest control time, with an SR similar to that of the visual memory designs. Appendix~\ref{sec:supp:design} details the protocol of this design study.

\paragraph{Outcome feedback.} The working memory retains most of its gain without terminal outcome feedback. Eight tasks test this signal: the four high-precision tasks and four multi-layout tasks, namely dishwasher loading, the fruit task, utensil placement, and dish placement. When the terminal outcome $r_k$ is withheld during online adaptation along their 20-scene chains, the overall SR decreases from 73.8\% to 65.0\%. The high-precision chains keep 52.5\% SR in both conditions. On the final five episodes of the four multi-layout chains, where task memory alone also runs on the same scenes, working memory without outcome feedback reaches 80.0\% SR, compared with 35.0\% for task memory alone and 95.0\% with outcome feedback, thereby retaining three quarters of the improvement through self-assessment of $\xi_k$ alone, as detailed in Appendix~\ref{sec:supp:outcome}.
\par\smallskip\noindent\colorbox{astralight}{\parbox{\dimexpr\linewidth-2\fboxsep\relax}{\textbf{Insight 1.} A bounded text working memory suffices: it matches the SR of visual memory at lower cost and keeps most of its gain without outcome feedback.}}\par

\begin{table}[t]
  \centering
  \footnotesize
  \setlength{\tabcolsep}{3pt}
  \caption{Working memory under the two agent harnesses of \astra on the chains and with the columns of Table~\ref{tab:design}. The visual-replay row includes text working memory, and the trajectory row extends it with the executed motion of $\pi_{0.5}$, as Appendix~\ref{sec:supp:trajectory} describes. Bold marks the best value within each harness.}
  \label{tab:harness}
  \newcolumntype{W}[1]{>{\hsize=#1\hsize\centering\arraybackslash}X}%
\begin{tabularx}{\linewidth}{l W{1.05} W{0.85} W{0.85} W{0.85} W{0.85} W{1.05} W{1.15} W{1.35}}
\toprule
\hd{Memory} & \hd{Coffee} & \hd{Gear} & \hd{Peg} & \hd{Nut} & \hd{SR} & \hd{Score} & \hd{Tokens} & \hd{Minutes} \\
\midrule
\rowcolor{tabband} \multicolumn{9}{c}{\textit{Harness with $\pi_{0.5}$}} \\
\midrule
Task memory only & 0.0 & \best{60.0} & 10.0 & \best{20.0} & 22.5 & 41.3 & \best{1.73} & 27.9 \\
+ text working memory & 0.0 & \best{60.0} & \best{30.0} & 10.0 & \best{25.0} & \best{51.6} & 2.88 & \best{26.5} \\
Text working memory + visual replay & \best{10.0} & 20.0 & 10.0 & 10.0 & 12.5 & 40.7 & 2.57 & 27.0 \\
Text + $\pi_{0.5}$ trajectory & \best{10.0} & 40.0 & 0.0 & \best{20.0} & 17.5 & 43.2 & 2.78 & 28.0 \\
\midrule
\rowcolor{tabband} \multicolumn{9}{c}{\textit{Direct control}} \\
\midrule
Task memory only & 0.0 & 40.0 & 10.0 & \best{10.0} & 15.0 & 40.5 & 3.51 & 29.5 \\
\rowcolor{astralight} + text working memory (\method) & \best{20.0} & \best{70.0} & \best{90.0} & \best{10.0} & \best{47.5} & \best{64.1} & \best{2.86} & \best{23.0} \\
\bottomrule
\end{tabularx}

\end{table}

\subsection{Memory Pays Off Under Direct Control}
\label{sec:analysis:delegation}
Table~\ref{tab:harness} shows a larger SR gain from the proposed memory mechanism under direct control than in the harness with $\pi_{0.5}$ on the same chains. Under direct control, peg insertion rises from one success in ten scenes to nine. In the harness with $\pi_{0.5}$, adding text working memory moves the SR from 22.5\% to 25.0\%, and adding the visual replay lowers it to 12.5\%.

Neither exposing the motion of the policy nor adding an external critic closes this gap. When the harness reports the executed end-effector trajectory of $\pi_{0.5}$ after each call and the MLLM reflects on it, the SR of the harness falls to 17.5\%, as Appendix~\ref{sec:supp:trajectory} details.

In an independent experiment, a critic-driven RSI loop lets an instance of \astra inspect training episodes over five rounds and revise a playbook and the harness with $\pi_{0.5}$. The loop raises the SR of this harness only from 44.4\% with an empty playbook to 46.7\% on 90 paired test episodes of 18 tasks. On the same episodes, \method under direct control reaches 76.7\% SR and matches or exceeds the critic-driven loop on 17 of the 18 tasks, with per-task results in Appendix~\ref{sec:supp:rsi}. Under direct control, reflection revises the low-level commands that the MLLM issues itself, whereas in the harness with $\pi_{0.5}$ it can only update the guidance for calling a fixed policy.
\par\smallskip\noindent\colorbox{astralight}{\parbox{\dimexpr\linewidth-2\fboxsep\relax}{\textbf{Insight 2.} Within the compared frameworks, the proposed memory evolution improves direct control more than the harness with policy.}}\par

\begin{figure*}[t]
  \centering
  \includegraphics[width=\linewidth]{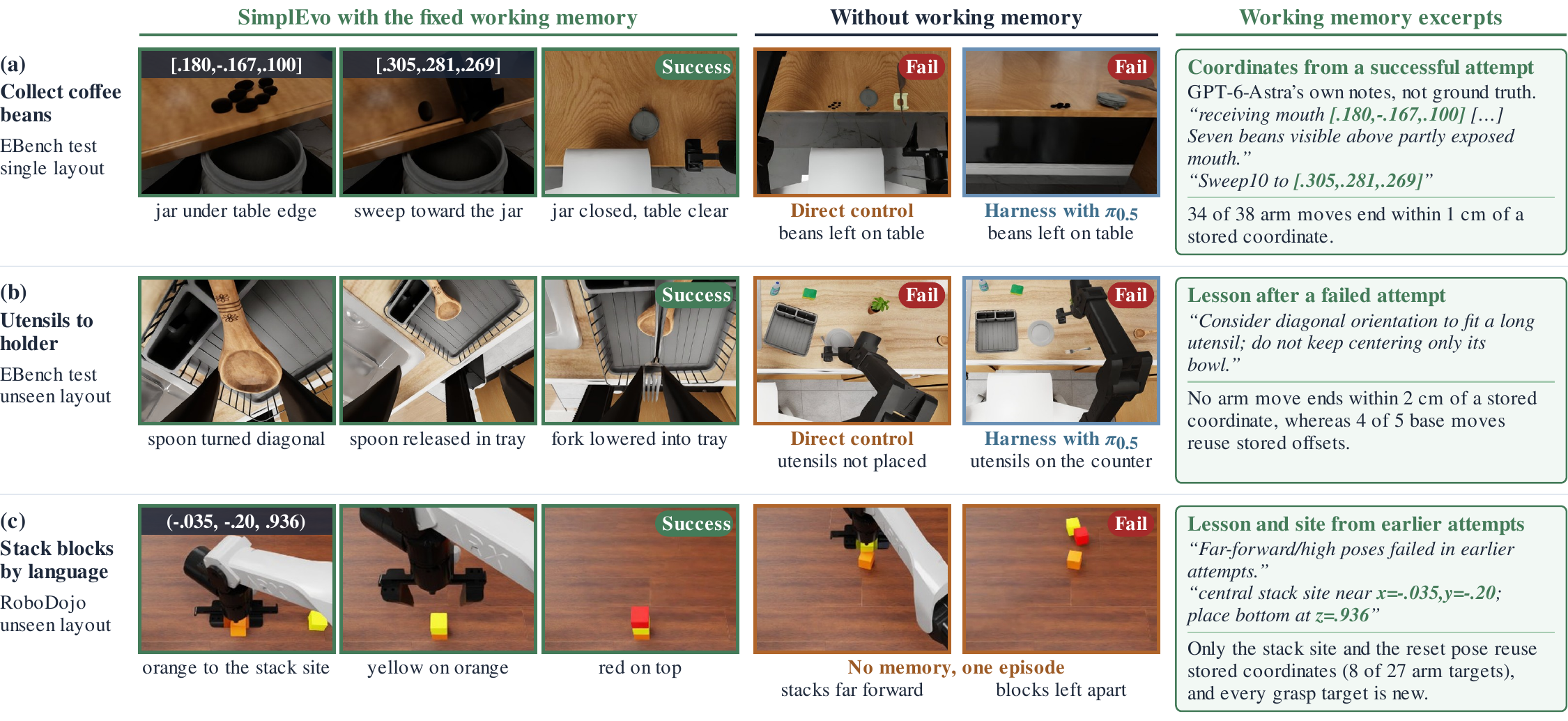}
  \caption{Test episodes with the fixed working memory of \method (green) and without working memory on the same scenes, with verbatim excerpts of the memory read in each \method episode ([\ldots] marks omissions). Each row shows a success of \method on a scene where the other configurations fail. Over all test episodes of these tasks, \method succeeds in 18 of 20 in (a), 12 of 20 in (b), and 49 of the 50 test layouts in (c), compared with none in (a) and 2 in (b) for direct control without working memory, and none and 3 for the harness with $\pi_{0.5}$. Dark labels give commanded positions in meters equal to coordinates in the working memory, which \astra wrote from its own commands and proprioception in training episodes without access to ground-truth object positions. Baseline frames in (a) and (b) come from Table~\ref{tab:ebench} and in (c) from Appendix~\ref{sec:supp:matched}.}
  \label{fig:cases}
\end{figure*}

\begin{table}[t]
  \centering
  \footnotesize
  \setlength{\tabcolsep}{3pt}
  \caption{SR (\%) by layout overlap on EBench and scene configuration on RoboDojo. Seen EBench layouts occur in at least one completed training episode of \method. The same partition applies to every configuration of \astra. On RoboDojo, each Generalization task pairs a standard configuration with a randomized one, the other rows are the entries of the public leaderboard (2 November 2026), and the 15 training layouts of \method include standard and randomized configurations. SR is averaged within each task and then across tasks, and $\Delta$ is the change from the first to the second column. Class marks vision-language-action models (VLA), world action models (WAM), and MLLM agents (Agent), and Agent + VLA or Agent + WAM denotes an agent that calls such a policy. Bold and underline mark the best and second-best values within each benchmark.}
  \label{tab:generalization}
  \begin{tabular*}{\linewidth}{@{\extracolsep{\fill}}llccc@{}}
\toprule
\hd{System} & \hd{Class} & \hd{Seen\,/\,standard} & \hd{Unseen\,/\,randomized} & \hd{$\Delta$} \\
\midrule
\rowcolor{tabband} \multicolumn{5}{c}{\textit{EBench, 15 multi-layout tasks: seen and unseen layouts}} \\
\midrule
Direct control~\citep{yao2026profiling} & Agent & 73.8 & 63.1 & $-$10.7 \\
Harness with $\pi_{0.5}$~\citep{yao2026profiling} & Agent + VLA & \second{77.2} & 69.6 & $-$7.6 \\
Harness with InternW0-$\Delta$~\citep{yao2026profiling} & Agent + WAM & 74.5 & \second{71.7} & \second{$-$2.8} \\
\rowcolor{astralight} \method (ours) & Agent & \best{83.1} & \best{87.5} & \best{$+$4.4} \\
\midrule
\rowcolor{tabband} \multicolumn{5}{c}{\textit{RoboDojo, 12 Generalization tasks: standard and randomized configurations}} \\
\midrule
Awomo-0.5~\citep{awomowam2026awomo} & WAM & 30.89 & 16.78 & $-$14.11 \\
VPP2~\citep{guo2026vpp2} & WAM & \second{46.44} & 12.56 & $-$33.88 \\
WALL-OSS-1.0-lite~\citep{zhai2025igniting} & VLA & 45.44 & 30.44 & $-$15.00 \\
\astra, zero-shot~\citep{zhang2026robodojo} & Agent & 32.67 & 28.33 & $-$4.34 \\
WorldScape-R$^2$SI~\citep{manifold2026r2si} & Agent + WAM & 40.33 & 31.33 & $-$9.00 \\
HexaShell-Preview~\citep{gao2026hexashell} & Agent & 38.00 & 31.00 & $-$7.00 \\
PyRUA-Lean~\citep{si2026pyrua} & Agent & 45.33 & \second{41.00} & \second{$-$4.33} \\
GO-3 Nano Preview & Agent + WAM & 43.00 & 24.33 & $-$18.67 \\
\rowcolor{astralight} \method (ours) & Agent & \best{52.00} & \best{52.00} & \best{0.00} \\
\bottomrule
\end{tabular*}

\end{table}

\subsection{How the Memory Generalizes to Unseen Scenes}
\label{sec:analysis:layouts}
\label{sec:results:why}
With its memories fixed, \method sustains its SR on unseen layouts and randomized configurations, whereas the SR of every other system in Table~\ref{tab:generalization} drops on them. On the 15 multi-layout EBench tasks with both kinds of test episodes, \method succeeds more often on unseen layouts than on seen ones, and each configuration of \astra without working memory drops by 2.8 to 10.7 points. On the 12 Generalization tasks of RoboDojo, \method keeps 52.00\% SR in both configurations, whereas the zero-shot entry of the same MLLM drops from 32.67\% to 28.33\% and the two strongest post-trained policies lose 15.00 and 33.88 points. Matched comparisons that rerun task memory alone on the same test scenes isolate the contribution of the working memory. With the task memory $\Mtask$ held fixed, adding the fixed working memory raises the SR from 67.4\% to 85.3\% on the last five test episodes of each of the 19 multi-layout EBench tasks and from 40.83\% to 66.11\% on three test layouts of each of the 42 RoboDojo tasks, as Appendix~\ref{sec:supp:matched} details.

The working memory pairs distilled lessons with evidence from earlier attempts, as in Figure~\ref{fig:cases}a, and the episode prompt of Appendix~\ref{sec:supp:protocol} presents these notes as hypotheses to verify against the current views. On an unseen layout, as in Figure~\ref{fig:cases}b and~c, the MLLM locates the objects in the current views and follows the procedural, completion, and recovery lessons, so the commanded positions of successful episodes rarely coincide with stored ones, and in the median episode only 11\% of the moving waypoints come within 1\,cm of any stored position. Appendices~\ref{sec:supp:replay} and~\ref{sec:supp:cases} analyze the commanded waypoints and present further case studies.
\par\smallskip\noindent\colorbox{astralight}{\parbox{\dimexpr\linewidth-2\fboxsep\relax}{\textbf{Insight 3.} Lessons about task invariants, such as procedure, completion checks, and error recovery, hold across layouts and scene conditions, so a memory built from them generalizes beyond the scenes that produced it.}}\par

\begin{table}[t]
  \centering
  \footnotesize
  \setlength{\tabcolsep}{2.5pt}
  \caption{Transfer of the working memory of \astra across MLLMs, measured on the 130 test episodes of the Mixed generalization setting of EBench, five per task across all 26 tasks: SR (\%) and mean score with task memory only and with the working memory of \astra read without modification. The working-memory entry of \astra comes from Table~\ref{tab:ebench}. Bold marks the higher value for each model.}
  \label{tab:transfer}
  \begin{tabularx}{\linewidth}{l*{4}{>{\centering\arraybackslash}X}}
    \toprule
    & \multicolumn{2}{c}{\hd{Task memory only}} & \multicolumn{2}{c}{\hd{Task + working memory of \astra}} \\
    \cmidrule(lr){2-3}\cmidrule(lr){4-5}
    \hd{Model} & \hd{SR} & \hd{Score} & \hd{SR} & \hd{Score} \\
    \midrule
    \rowcolor{tabband} \multicolumn{5}{c}{\textit{Other MLLMs}} \\
    \midrule
    Gemini 3.8 Flash & 23.8 & 37.8 & \best{53.8} & \best{71.7} \\
    DeepSeek-V4.1-Flash & 0.8 & 4.6 & \best{13.1} & \best{26.5} \\
    \midrule
    \rowcolor{tabband} \multicolumn{5}{c}{\textit{MLLM that wrote the working memory}} \\
    \midrule
    \astra & 60.8 & 76.8 & \best{81.5} & \best{89.3} \\
    \bottomrule
  \end{tabularx}
\end{table}

\subsection{Memory Transfers Across Models}
\label{sec:analysis:transfer}
Read without modification, the fixed working memory of \astra improves other MLLMs under direct control on the Mixed generalization setting of EBench across all 26 tasks, as shown in Table~\ref{tab:transfer}. Each MLLM attempts every scene of this dimension in two independent sessions, once with task memory alone and once with the memory that \astra uses in Table~\ref{tab:ebench}. With this memory, Gemini 3.8 Flash more than doubles its SR, DeepSeek-V4.1-Flash succeeds on scenes it almost never solves with task memory alone, and \astra itself improves as well. Appendix~\ref{sec:supp:transfer} breaks these gains down by task layout and extends the study to four MLLMs along the adaptation chains of two tasks. Transfer remains limited when the working memory omits low-level control behaviors. Over 24 paired episodes of eight tasks, both with and without the memory, DeepSeek-V4.1-Flash sets the gripper target near the measured width of the object in 19\% of its lifting waypoints and thus does not squeeze the object. \astra does so in only 5\% of its lifting waypoints because it consistently closes the gripper fully on a lifted object, and its working memory therefore never states this low-level control habit.
\par\smallskip\noindent\colorbox{astralight}{\parbox{\dimexpr\linewidth-2\fboxsep\relax}{\textbf{Insight 4.} Text working memory carries the explicit execution lessons of a stronger MLLM to weaker MLLMs without weight updates, whereas low-level control habits that the memory leaves unstated do not transfer.}}\par

\section{Conclusion}
\label{sec:conclusion}

Although embodied agents increasingly surround frontier MLLMs with perception pipelines, harnesses with policy, and critic loops, our results show that a frontier MLLM under direct control already achieves strong closed-loop manipulation once its within-episode corrections persist across episodes. \method retains these corrections in two self-written text memories: a coordinate-free task memory distilled once from one training demonstration, and a bounded working memory rewritten after each training episode. Under the strict train--test protocol, with RGB views and proprioception as its only observations, \method raises \astra from 46.7\% to 81.7\% SR on EBench and leads the RoboDojo leaderboard. Our analyses yield four takeaways for embodied agent design. (1) A bounded text memory distilled from the trajectories of the agent itself matches the SR of visual memory at lower cost and retains most of the improvement without outcome feedback. (2) The proposed memory mechanism improves direct control more than the compared harness with policy. (3) Lessons on task invariants, applied to the objects that the MLLM locates in the current views, hold on unseen layouts and under randomized configurations. (4) A text memory written by one MLLM also improves other MLLMs that read it without modification, though only for the behaviors that it states explicitly.

\paragraph{Limitations and future directions.} These findings point to three directions for future work. First, \emph{verified memory updates} could validate each candidate rewrite before it enters the working memory and thus filter out spurious lessons. Second, \emph{object-centric, behavior-explicit memory} could anchor recorded actions to visual features and state the low-level control habits that \astra leaves implicit, so that recorded evidence carries across workspace geometries and the memory serves weaker MLLMs. Third, both benchmarks pause the simulator during MLLM deliberation and end episodes automatically upon completion, and each episode costs millions of tokens. These conditions motivate \emph{asynchronous, token-efficient deployment on physical robots} with autonomous completion detection~\citep{openrua2026revisiting}.

\bibliographystyle{iclr2027_conference}
\bibliography{arxiv/arxiv}

\clearpage
\appendix
\counterwithin{figure}{section}
\counterwithin{table}{section}
\renewcommand{\topfraction}{0.9}\renewcommand{\bottomfraction}{0.8}\renewcommand{\textfraction}{0.07}\renewcommand{\floatpagefraction}{0.8}
\setcounter{topnumber}{3}\setcounter{bottomnumber}{2}\setcounter{totalnumber}{5}
\let\appxsection\section\renewcommand{\section}{\FloatBarrier\appxsection}
\let\appxsubsection\subsection\renewcommand{\subsection}{\FloatBarrier\appxsubsection}
\section*{Appendix}
\makeatletter
\let\appxacl\addcontentsline
\renewcommand{\addcontentsline}[3]{\ifstrequal{#1}{toc}{\appxacl{atc}{#2}{#3}}{\appxacl{#1}{#2}{#3}}}
{\setcounter{tocdepth}{2}\@starttoc{atc}}
\makeatother
\clearpage
\newcommand{\fitwidth}[1]{\setbox0\hbox{#1\unskip}\par\noindent\makebox[\linewidth][c]{\ifdim\wd0>\linewidth\resizebox{\linewidth}{!}{\box0}\else\box0\fi}\par}

\newenvironment{promptquote}{\begin{list}{}{\leftmargin=0.8em\rightmargin=0pt\topsep=4pt\partopsep=0pt\parsep=3pt\itemsep=0pt}\item\relax\scriptsize\ttfamily\raggedright}{\end{list}}

\section{Information Protocol and Agent Interfaces}
\label{sec:supp:protocol}
This section summarizes in Table~\ref{tab:supp:protocol} the inputs and actions of each configuration of \astra under the protocol of Section~\ref{sec:method:setting} and reproduces the instructions and tool descriptions of the agent harness verbatim. The configurations differ in the tools exposed to the agent and in the memory retained across episodes.

\begin{table}[!htbp]
  \centering
  \footnotesize
  \setlength{\tabcolsep}{3pt}
  \caption{Inputs and actions of the configurations of \astra on EBench. Proprioception covers the end-effector poses, the grippers, and the mobile base. The demonstration is one training recording per task, read through its annotation and the read-only demonstration tools. Outcomes are the terminal success and score of earlier adaptation episodes, and only \method receives them.}
  \label{tab:supp:protocol}
  \fitwidth{\begin{tabular}{@{}lccc@{}}
    \toprule
    & Direct & Harness with & \\
    Input or action & control & policy & \method \\
    \midrule
    RGB views (overview, two wrists) & \checkmark & \checkmark & \checkmark \\
    Proprioception & \checkmark & \checkmark & \checkmark \\
    Instruction, robot constants & \checkmark & \checkmark & \checkmark \\
    Demonstration, annotation & \checkmark & \checkmark & \checkmark \\
    End-effector waypoints & \checkmark & \checkmark & \checkmark \\
    Calls to a post-trained policy & -- & \checkmark & -- \\
    Working memory & -- & -- & \checkmark \\
    Outcomes of earlier episodes & -- & -- & adaptation \\
    \bottomrule
  \end{tabular}}
\end{table}

\subsection{Interface on EBench}
\label{sec:supp:protocol:ebench}
Every EBench configuration of \astra receives the same base system instructions. Direct control appends the last sentence of the following prompt, \emph{Choose continuous poses, motion timing and observation cadence yourself}, and the harnesses with policy append the description of their policy tool in its place.

\begin{promptquote}
You control one simulated EBench robot episode. Use only the supplied RGB and measured robot state and the declared robot tools. Do not read external files, use shell, browse, or delegate to other agents. The server judges success independently. Stop when it reports a terminal result or a budget is exhausted. Repeated observe at the same timestep shows the same scene. A closed gripper or finger contact alone does not prove a secure grasp. Choose continuous poses, motion timing and observation cadence yourself.
\end{promptquote}

The first message of an episode states the instruction, the robot specifications, and the resource limits, followed by guidance on grasp verification that all configurations share and by the paragraphs on the demonstration and the working memory. The message below opens a test episode of peg insertion with the fixed working memory.

\begin{promptquote}
Complete this single task yourself using the tools: Lift the peg and the hole at the same time, put the peg into the hole, and then set them back on the table.

Robot: [\ldots] two arms and a movable planar base. Left arm is left in the overlook view. EEF poses are local to each arm's planner base: +X forward, +Y robot-left, +Z up; quaternion order wxyz. Fingers extend along tool +X and open along tool Y. Wrist cameras move/rotate with the hands. Measured local EEF z is not height above the tabletop. There is no supplied depth, object pose or camera calibration. Only RGB and measured robot EEF/gripper/base state are available. Base observed yaw is radians; tool command yaw is degrees. You have one continuous conversation and may choose any continuous EEF target, both arms, base motion and waypoint batching. Use execute\_eef sequentially, not parallel calls. Use observations to judge motion and object manipulation. The server determines task success independently. Do not start another episode. Resource limits: unlimited tool calls, physics deadline enforced by the server; no hard token limit; [\ldots] Select only useful camera views after the initial observation to control resource use. When finished, state the observed outcome concisely.

Manipulation guidance: after visually aligning, close at a stationary pose and allow settling. A nonzero finger gap on the table is contact evidence, not proof of a secure grasp. Test each grasp by lifting only 0.01 m first, using motion\_steps=16 or more, then hold 8..16 steps and request the active wrist plus overlook views. Keep orientation and base fixed during this test. Check both finger gap and RGB to confirm the object rises with the gripper before further lifting or transport. If it remains on the table or slips, lower and adjust local contact depth/centering using current images. There is no fixed limit on failed grasp attempts; retry or change approach within the remaining experiment budgets. Avoid large instantaneous target jumps while holding an object; use motion\_steps for controlled transport. Total steps per waypoint must remain \textless{}=64 and includes the ramp. Use only the active wrist for local corrections, adding overlook when establishing alignment, testing grasp, or verifying placement.

RGB inputs are downsampled: overlook 640x360, wrist 384x288; no crop or object annotation.

The saved ICL demonstration remains available through query\_icl throughout this episode. Revisit it when the target region, stage order or arm roles are unclear, and when repeated failures or a change of approach make the reference useful. Use view=text for the original annotation and keyframe index, view=frames with frame\_indices for selected images, or view=all for the full reference. Indices are zero-based entries in the keyframe index, not source video frame numbers. Querying does not advance simulation. Ground actions in the current RGB/state.

Working memory: read\_task\_notes holds notes about this task recorded by earlier attempts at it in other scenes (it may be empty); they are read-only in this attempt. read\_working/write\_working is this attempt's scratchpad; history, inspect and read\_state\_history give exact records and archived views of this attempt; each tool result states its call number. Before each robot action, briefly predict what you expect to see; after it, describe what actually changed. Notes are hypotheses: verify them against the current views.
\end{promptquote}

The agent commands the robot through three tools. The tool \texttt{execute\_eef}, whose description follows, executes all arm and base actions. The tool \texttt{set\_gripper} commands a normalized gripper opening while both arms and the base remain stationary, and \texttt{observe} returns the current camera views and robot state without advancing the simulation.

\begin{promptquote}
Execute 1..6 native GenManip absolute EEF waypoints in order. Each arm target has xyz(m), unit quaternion(wxyz) in that arm's planner-base frame, and two finger joint targets(m): [0.044,0.044] open, [0,0] close. No translation or angle quantization. Both arms and the mobile base may move together. Null arm holds its measured local pose and prior gripper target; null base holds measured planar base pose. Base x/y are metres; command yaw is DEGREES, while observed state.base yaw is RADIANS. Each waypoint consumes 'steps' physics steps (1..64). Optional motion\_steps (0..steps, default 0) linearly ramps position and spherically interpolates rotation from measured pose to target over that many physics steps, then holds for the remainder. Base pose also ramps. Finger commands apply immediately, so close in a separate stationary waypoint. With motion\_steps=0 the full target applies immediately: extra steps are holding time, not slow motion. All commands are absolute; there is no repeated relative displacement. The simulator performs IK; targets may remain unreachable, so inspect returned measured poses. No automatic grasp approval or task-stage constraints are imposed. Returns final robot state and selected RGB cameras; sequences have no intermediate model observation. You choose target poses, timing, observation cadence, camera subset and when to batch waypoints.
\end{promptquote}

\subsection{Task Memory Annotation}
\label{sec:supp:protocol:annotation}
The annotator is an instance of \astra at high reasoning effort with three read-only tools. The tool \texttt{demo\_search} divides the recording into action segments and gripper events and reports, for each segment, the moving arm or base, its displacement, and the measured gripper opening. The tool \texttt{demo\_inspect} returns frames from the overview and wrist cameras, and \texttt{demo\_actions} reports the measured arm and base motion over a frame range relative to its first frame. The annotator receives the following message, here for peg insertion, and submits the task memory once.

\begin{promptquote}
You are annotating one recorded robot demonstration so that a robot agent can learn from it later. The robot is a mobile bimanual manipulator; the demonstration was recorded in simulation. Task instruction: "Pick up the peg and the hole simultaneously, insert the peg into the hole, and place them back on the table.". The demonstration has 1943 frames (one frame is one execution step).

Use demo\_search to list its motion bouts and gripper events, demo\_inspect to look at frames from the overlook, left\_wrist and right\_wrist cameras, and demo\_actions to read measured arm motion. Review the WHOLE demonstration, including its last frames: stages are easy to miss when a scene is dark or an arm moves without a gripper event (for example closing a door or a lid). Then call submit\_icl once.

Write for a robot agent that will face a different layout of the same task: describe operations, arm roles, order, contacts and the visible end state; say what you saw versus what you infer; do not report absolute coordinates or frame-specific numbers as instructions. There is no success label for this recording; the instruction defines what must be true at the end. [\ldots]
\end{promptquote}

\subsection{Working Memory Tools and Reflection}
\label{sec:supp:reflection}
During an episode, the MLLM reads the working memory through \texttt{read\_task\_notes}. It also keeps a scratchpad for the current episode (\texttt{read\_working} and \texttt{write\_working}), lists its tool calls (\texttt{history}), revisits frames it received (\texttt{inspect}), and reads the measured robot states of the episode (\texttt{read\_state\_history}). It can also revisit the demonstration through \texttt{query\_icl}, which returns the annotation and selected keyframes, and through the three demonstration tools of the annotator. During adaptation, the MLLM rewrites the working memory through \texttt{write\_task\_notes}, whose description follows.

\begin{promptquote}
Replace your task notes (at most 32768 bytes UTF-8; nothing is truncated). [\ldots] Write reusable, revisable, task-specific knowledge: what worked, what failed and why, with the evidence (attempt and call numbers). Coordinates measured in one scene are not commands for another scene.
\end{promptquote}

After each adaptation episode, the agent receives one reflection turn in which only the memory tools are available. The turn opens with the following message, and the variant without outcome feedback of Appendix~\ref{sec:supp:outcome} replaces the result with a statement that the official result is not shown.

\begin{promptquote}
The episode has ended. Official result from the evaluation server: success=\textless{}SR\textgreater{}, score=\textless{}Score\textgreater{}. No robot action is possible now; only the memory tools work. Review this attempt if useful, then update your task notes with write\_task\_notes (at most 2 writes) so that later attempts at this task in other scenes do better: what worked, what failed and the evidence (attempt and call numbers). Keep the notes reusable and revisable. Reply with one short sentence when done.
\end{promptquote}

\subsection{Interface on RoboDojo}
\label{sec:supp:protocol:robodojo}
The RoboDojo agent harness retains the tools and the message structure of EBench, with the robot paragraph adapted to the fixed bimanual tabletop setup.

\begin{promptquote}
Robot: two ARX X5 6-DoF arms with parallel grippers on a fixed table (no mobile base). Shared world frame (metres): the robot faces +Y; the left arm's base is at (-0.3, -0.45, 0.765) and the right arm's at (0.3, -0.45, 0.765), so +X is toward the robot's right and +Z is up; the tabletop is at z = 0.765. EEF poses are the arm flange (link6) pose in this frame, quaternion order wxyz. The fingers extend along the flange's +X axis and the jaws open and close along its Y axis; the centre of the finger pads is about 0.15 m from the flange origin along +X. For example, quaternion (0.5, -0.5, 0.5, 0.5) points the fingers straight down with the jaws closing along world X. Each episode starts with both arms at the same home pose (all joints zero, grippers open). Cameras: cam\_head (fixed above and behind the arms, looking down at the table), cam\_left\_wrist and cam\_right\_wrist (move and rotate with the hands). There is no supplied depth, object pose or camera calibration. Only RGB and measured EEF/gripper state are available; the gripper value is your last command (1 open, 0 closed), not a measured finger width.
\end{promptquote}

\section{Memory Examples}
\label{sec:supp:memory}
This section presents the two memories of \method introduced in Sections~\ref{sec:method:task} and~\ref{sec:method:working} as the agent reads them and traces how the working memory develops along the adaptation chains. All excerpts are quoted verbatim, and [\ldots] marks omitted text.

\subsection{Task Memory}
\label{sec:supp:taskmem}
Figure~\ref{fig:supp:taskmem} reproduces the task memory of peg insertion on EBench. It describes the arm roles and 11 stages in order and closes with the observed outcome. Like the task memories of the other EBench tasks and the 34 demonstration-equipped tasks of RoboDojo, it contains no coordinates. In every episode of the task, the MLLM reads this text together with 15 keyframes of the recording, each with a one-line caption.

\begin{figure*}[!htbp]
  \centering
  \noindent\colorbox{astralight}{\parbox{\dimexpr\linewidth-2\fboxsep\relax}{\scriptsize\raggedright One recorded demonstration of this task. Reference instruction: Pick up the peg and the hole simultaneously, insert the peg into the hole, and place them back on the table. \par Arm roles: The left arm grasps the outside of the receiving hole block, lifts and positions it with its opening accessible, holds it nearly stationary during insertion, and finally places the assembled block and peg on the table. The right arm grasps the yellow peg, extracts it from a separate tabletop holder, aligns and inserts it into the left-held block, releases it, and withdraws. The base remains essentially stationary. The observed pickups are sequential, with the hole picked up first; both objects are held concurrently during alignment and insertion. \par Observed sequence: Initially a receiving block with an empty central hole and a separate holder containing an upright yellow peg rest on the table. The left gripper approaches, opens around the receiving block, closes against its outer sides, and lifts and shifts the block toward the shared workspace. While the left arm holds the hole accessible, the right gripper approaches and closes around the exposed peg. The right arm lifts the peg out, leaving its original holder on the table. It carries the peg toward the left-held opening and adjusts its orientation and lateral position to bring the lower end into alignment. With the left block steady, the right arm gradually advances the peg into the central hole, making small pose corrections. The right gripper opens, leaving the peg protruding from and supported by the receiving block, then lifts away and retracts. The left arm lowers the assembled pair onto the tabletop, opens its gripper, lifts clear, and returns to rest. The final view shows the peg still in the receiving block, the original holder empty nearby, and both arms withdrawn. \par Stages (frame = one execution step of the recording): \par - frames 0-25 (none; seen): Initial tabletop arrangement. Both hands begin clear of the objects. An empty receiving block and a separate holder containing the yellow peg sit on the tabletop. \par - frames 25-192 (left; seen): Approach and surround the receiving block. The left wrist turns and approaches the receiving block, opens its fingers, and lowers them to opposite outer sides while leaving the central hole unobstructed. \par - frames 192-354 (left; seen): Grasp, lift, and hold the receiving block. The left fingers close on the block; the arm lifts and shifts it toward the assembly workspace, then holds the opening accessible. The right hand remains clear. \par - frames 354-562 (right; seen): Approach the peg while holding the hole. The right arm approaches the yellow peg, opens, and lowers its fingers around the exposed shaft. The left arm maintains its grasp on the receiving block. \par - frames 562-832 (right; seen): Grasp and extract the peg. The right fingers close around the peg and hold it while the arm gradually lifts it out of its original holder. A clear gap appears between the peg tip and the now-empty holder. The left hand continues holding the receiving block. \par - frames 832-1100 (right; seen): Carry and align the peg over the receiving hole. The right arm brings the free peg toward the left-held block, changing wrist orientation and lateral position so its lower end approaches the central opening. The left arm remains nearly stationary. \par - frames 1100-1479 (right; seen): Advance the peg into the held hole. The right arm lowers and adjusts the peg at the opening while keeping its grasp. The shaft becomes visibly engaged in the receiving hole. Small alignment and insertion corrections continue with the left gripper supporting the block; the exact contact onset and seating depth are partly occluded. \par - frames 1479-1679 (right; seen): Release the inserted peg and withdraw the right hand. The right fingers open, leaving the peg standing in the receiving block. The right hand rises clear, closes empty during retraction, and returns to its resting region. The left hand holds the assembled pair. \par - frames 1679-1822 (left; seen): Lower the assembled pair to the table. The left arm lowers the receiving block with the peg still inserted, placing its base against the tabletop and holding it briefly before release. \par - frames 1822-1855 (left; seen): Release the assembly and lift clear. The left fingers open away from the block, and the hand lifts clear. The peg and receiving block remain together on the table. \par - frames 1855-1943 (left; seen): Retract the left arm and leave the final arrangement. The left gripper closes empty as the arm retracts to rest. Through the last frame, both hands remain clear, the peg stays inserted in the receiving block on the table, and the original empty holder remains nearby. \par Observed outcome: The yellow peg is visibly inserted in the receiving block and remains there after both grippers release. The receiving block and peg rest together on the table at the end; the separate original peg holder remains empty on the table. Both arms are clear and their empty grippers are closed. This supports the insertion-and-return end condition. [\ldots] \par This is a separate recorded episode with no success label; the live instruction defines what must be true at the end. Transfer operations, order and checks; ground every command in the live views and measured state.}}
  \caption{Task memory of peg insertion on EBench, written once by the annotator from one training recording and read in every training and test episode of the task.}
  \label{fig:supp:taskmem}
\end{figure*}

\subsection{Working Memory}
\label{sec:supp:workmem}
Figure~\ref{fig:supp:workmem} shows excerpts of two fixed working memories. On EBench peg insertion, a chain of nine training episodes with five successes wrote the memory that all 20 test episodes read, and one of its lessons warns that a single camera view cannot confirm that the peg is aligned with the hole. On RoboDojo stack blocks, all 15 training episodes succeeded, and one lesson prefers stacking on an existing reachable base over moving the base to a central site.

\begin{figure*}[!htbp]
  \centering
  \begin{minipage}[t]{0.495\linewidth}
  {\footnotesize\textbf{(a) EBench, peg insertion}}\par\smallskip
  \noindent\colorbox{astralight}{\parbox{\dimexpr\linewidth-2\fboxsep\relax}{\scriptsize\raggedright Task: pick up receiving block and peg, insert while both held, return assembly to table. Attempt 2 official success=1.0, score=1.0; terminal at call 44, timestep $\sim$1311 during lowering assembled pair. Attempt 1 failed at 2500 steps (score .5). Preserve physics for placement; unlimited calls are not unlimited physics. Coordinates below are scene-specific evidence, not commands to replay. \par[\ldots]\par - Left calls 4-7: open approach above receiving block; lower around outer sides; stationary close 24 steps; 1 cm test lift with 20 ramp+12 hold. \textcolor{astrablue}{Pose at grasp [.2807,-.0402,.0468], q [.8398,.1282,.2068,-.4854].} Finger joints $\sim$.014 each. Block stayed fixed in wrist image and visibly lifted in overlook. \par[\ldots]\par - Crucial: single-view tip overlap above hole is not 3D coaxial alignment. Peg/hand may maintain apparent position until rim contact, then slide/tip. At first stationary tip, changed cap attitude, or changing finger gap, relieve pressure promptly; do NOT add 1 cm or more descent. Use 1-3 mm corrections near contact. \par[\ldots]\par \textcolor{astrablue}{- Call16 applied Exp([22.52,-20.14,35.96]deg) to current right pickup quaternion, yielding q [.82513055,-.04968560,.12816274,.54796854].} \textcolor{astrablue}{Moved right [.310,.156,.185],48ramp+8hold.} Achieved q verified. This turn remained FIXED for all subsequent successful insertion commands. \par[\ldots]\par \textcolor{astrablue}{- Calls25-28: FIX XY [.308,.138] and q; descend z .121,.117,.113,.109 in 4mm increments, each16ramp+8hold, both wrists.} \par[\ldots]\par 2. Single left-view tip overlap can occur while the peg is still centimeters ABOVE the opening. Right wrist was useful after wrist reorientation: track tip-to-hole separation in BOTH views during descent. \par[\ldots]\par Updated reusable inference: Near-identical pickup/turn EEF poses can yield a different peg-in-hand pose, as attempts6/8/9 demonstrate. Stable gap$\sim$.00360 can be successful (attempt9), whereas a contact-induced decrease from an established larger gap warns of sliding. Absolute gap value alone is not the criterion. Use both wrist views, stable relative motion, and small live XY corrections; gradual loosening after promising engagement can reveal support.}}
  \end{minipage}\hfill
  \begin{minipage}[t]{0.495\linewidth}
  {\footnotesize\textbf{(b) RoboDojo, stack blocks}}\par\smallskip
  \noindent\colorbox{astralight}{\parbox{\dimexpr\linewidth-2\fboxsep\relax}{\scriptsize\raggedright GENERAL GUIDANCE: Attempts1-15 officially succeeded. Prefer two transfers onto an existing reachable base; three-transfer central relocation is robust but often takes $\sim$500steps. Verify reach, size, and alignment from current RGB, not historical coordinates. \textcolor{astrablue}{Standard down q=[.5,-.5,.5,.5], typical pickup flange z=.9255.} \par[\ldots]\par Keep orientation fixed while holding. Stationary close10 then mandatory1cm lift6ramp+4hold, head+wrist: confirm object moves with fingers, not just finger contact/command. \par[\ldots]\par WRIST: image right=world+X; moving hand-worldY moves object upward. Supporting cube appears ABOVE held cube when aligned, not overlapping image centers. \par[\ldots]\par Prefer a central base from outset when blocks span both sides; this would save the failed transport and recovery time. Reachability boundary observed here is pose-specific, not a universal X limit. \par[\ldots]\par \textcolor{astrablue}{Base released at flange .928, middle .963, top .998; all used [0,-.21].} \par[\ldots]\par Keep tilt FIXED throughout transfer so the cube stays upright; rotating gripper back to downward while holding would rotate the cube. \textcolor{astrablue}{Call18 raised/retracted to[-.08,-.12,1.06] (18 ramp+2 hold), then moved to[-.26,-.062,1.07] (22+2).} Offset flange Y to account for forward fingers, instead of reusing downward-grasp flange XY. \par[\ldots]\par LOWERING solved high reach limitation; no need to repeat unreachable height or regrasp. Here .07m clearance left fingertips visibly above tower before retreat. Retain geometry/RGB checks, not coordinates.}}
  \end{minipage}
  \caption{Excerpts of two fixed working memories of \astra: (a) peg insertion on EBench and (b) stack blocks on RoboDojo. Blue marks sentences that contain a stored pose (a position in meters or an orientation quaternion), and black marks the remaining sentences.}
  \label{fig:supp:workmem}
\end{figure*}

\subsection{Chain Dynamics}
\label{sec:supp:chains}
On EBench, the 26 adaptation chains comprise 192 training episodes with 101 successes. Of these chains, 21 stop after three consecutive successes, 11 of them without a failure, and the other five reach the limit of 15 episodes. A training episode consumes 4.09M tokens on average, including its reflection turn.

\begin{figure}[!htbp]
  \centering
  \includegraphics[width=\linewidth]{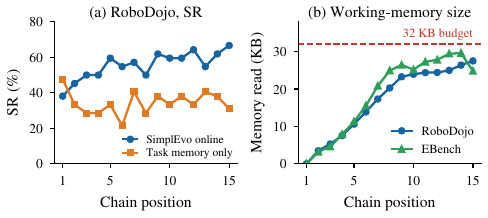}
  \caption{Adaptation chains of \method. (a) SR at each chain position of the online RoboDojo chains (42 episodes per point, averaged over tasks without dimension weighting) with working memory and with task memory only. (b) Mean size of the working memory read at each position, over the 42 RoboDojo chains and over the EBench adaptation chains still running at that position (26 at positions 1 to 3, 5 at positions 14 and 15).}
  \label{fig:supp:chain}
\end{figure}

\begin{table}[!htbp]
  \centering
  \footnotesize
  \setlength{\tabcolsep}{2.5pt}
  \caption{Online RoboDojo chains by blocks of five positions (210 episodes per cell): score (0--100) and mean tokens per episode for task memory only and for \method online, and the mean size of the working memory read (KB). Scores average episodes without dimension weighting, and Figure~\ref{fig:supp:chain}a gives the SR.}
  \label{tab:supp:chain}
  \fitwidth{\begin{tabular}{@{}lccccc@{}}
    \toprule
    & \multicolumn{2}{c}{Task memory only} & \multicolumn{3}{c}{\method online} \\
    \cmidrule(lr){2-3}\cmidrule(l){4-6}
    Positions & Score & Tokens & Score & Tokens & Memory \\
    \midrule
    1--5 & 41.55 & 1.42M & 55.93 & 1.65M & 5.4 \\
    6--10 & 39.64 & 1.42M & 62.88 & 1.75M & 19.7 \\
    11--15 & 42.76 & 1.41M & 67.60 & 1.77M & 25.6 \\
    \bottomrule
  \end{tabular}}
\end{table}

On RoboDojo, averaging Figure~\ref{fig:supp:chain}a over blocks of five positions, the online SR rises from 48.57\% in the first block to 61.43\% in the last, while task memory alone stays between 32.38\% and 36.19\%. Table~\ref{tab:supp:chain} gives the score, the token count per episode, and the memory size of each block. The working memory grows by about 3\,KB per episode over the first nine positions and then levels off below the budget, and the reflection turn adds 0.21M tokens per episode on average.

\section{Per-Task Results on EBench}
\label{sec:supp:ebench}
\begin{table*}[!htbp]
  \centering
  \footnotesize
  \setlength{\tabcolsep}{3.2pt}
  \caption{Per-task results on the 510 test episodes of EBench. Precision is low (L), medium (M), or high (H), and the horizon is short (S) or long (L). Train gives the successes and episodes of the adaptation chain, which stops after three consecutive successes or 15 episodes and yields the fixed working memory. The SR columns of \astra give the configurations without working memory in Table~\ref{tab:ebench} under direct control and in the harnesses with $\pi_{0.5}$ and InternW0-$\Delta$ (IW0-$\Delta$). Replay is the mean percentage of moving waypoints in Appendix~\ref{sec:supp:replay} that come within 1\,cm of a coordinate in the fixed working memory, and a dash marks a memory that stores no coordinates. Seen\,/\,unseen gives the successes of \method on test episodes whose layout appears, or does not appear, among the training episodes of the chain, and a dash marks single-layout tasks or a split without test episodes. The last row gives totals for Train and $n$ and means over tasks in the other columns. Bold marks the best SR in each row.}
  \label{tab:supp:ebench_pertask}
  \fitwidth{\begin{tabular}{@{}lcccrrrrrrrrr@{}}
\toprule
 & & & & & & \multicolumn{3}{c}{SR (\%), \astra} & \multicolumn{4}{c}{\method} \\
\cmidrule(lr){7-9}\cmidrule(l){10-13}
Task & Prec. & Hor. & Layout & Train & $n$ & Direct & $\pi_{0.5}$ & IW0-$\Delta$ & SR & Score & Replay & Seen\,/\,unseen \\
\midrule
apple from shelf & L & S & multi & 3/3 & 20 & 70.0 & \textbf{85.0} & 75.0 & \textbf{85.0} & 85.0 & -- & 9/9\,/\,8/11 \\
apple to fruit bowl & L & S & multi & 5/6 & 20 & 90.0 & 95.0 & \textbf{100.0} & \textbf{100.0} & 100.0 & 7 & 8/8\,/\,12/12 \\
bookmark on book & L & S & multi & 3/3 & 20 & \textbf{90.0} & 85.0 & \textbf{90.0} & \textbf{90.0} & 90.0 & 34 & 3/3\,/\,15/17 \\
bottle & M & L & multi & 3/13 & 20 & 0.0 & 0.0 & 0.0 & \textbf{45.0} & 88.8 & 27 & 5/10\,/\,4/10 \\
bowl to plate & L & S & multi & 3/3 & 20 & 75.0 & \textbf{95.0} & \textbf{95.0} & 80.0 & 80.0 & 19 & 3/5\,/\,13/15 \\
collect coffee beans & H & L & single & 5/13 & 20 & 0.0 & 0.0 & 0.0 & \textbf{90.0} & 95.0 & 89 & -- \\
detergent & L & L & multi & 3/3 & 20 & 55.0 & \textbf{90.0} & 70.0 & \textbf{90.0} & 96.7 & 16 & 1/2\,/\,17/18 \\
dish & M & L & multi & 4/5 & 20 & 50.0 & 60.0 & 70.0 & \textbf{90.0} & 98.0 & -- & 4/4\,/\,14/16 \\
dishwasher & L & L & multi & 3/3 & 20 & 5.0 & 60.0 & 75.0 & \textbf{85.0} & 94.2 & 26 & --\,/\,17/20 \\
flip cup collect cookies & M & L & single & 4/15 & 20 & 10.0 & 40.0 & \textbf{60.0} & 45.0 & 80.0 & 39 & -- \\
frame against pen holder & L & L & single & 3/3 & 20 & 65.0 & 65.0 & 45.0 & \textbf{100.0} & 100.0 & 57 & -- \\
fruit & L & L & multi & 3/3 & 20 & 40.0 & 30.0 & 45.0 & \textbf{95.0} & 98.0 & 3 & --\,/\,19/20 \\
install gear & H & S & single & 3/15 & 20 & 15.0 & 10.0 & 40.0 & \textbf{85.0} & 87.5 & 76 & -- \\
make sandwich & M & L & multi & 4/10 & 15 & 46.7 & 73.3 & \textbf{86.7} & 73.3 & 93.3 & 35 & 8/12\,/\,3/3 \\
microwave & L & L & multi & 5/6 & 15 & 73.3 & 60.0 & 60.0 & \textbf{86.7} & 97.8 & 71 & 3/3\,/\,10/12 \\
peg in hole & H & L & single & 5/9 & 20 & 20.0 & 5.0 & 0.0 & \textbf{85.0} & 92.5 & 69 & -- \\
pen & M & L & multi & 7/15 & 20 & 15.0 & 5.0 & 0.0 & \textbf{25.0} & 70.0 & 1 & 5/20\,/\,-- \\
perfume to cosmetics rack & L & S & multi & 3/5 & 20 & 75.0 & 80.0 & 80.0 & \textbf{95.0} & 97.5 & 29 & 4/5\,/\,15/15 \\
put glass in glassbox & M & L & single & 3/6 & 20 & 20.0 & \textbf{85.0} & 80.0 & 70.0 & 78.3 & 63 & -- \\
remote to holder & L & S & multi & 3/3 & 20 & \textbf{100.0} & 90.0 & \textbf{100.0} & \textbf{100.0} & 100.0 & 5 & 4/4\,/\,16/16 \\
salt to spice rack & L & S & multi & 3/3 & 20 & \textbf{100.0} & \textbf{100.0} & \textbf{100.0} & \textbf{100.0} & 100.0 & 7 & 3/3\,/\,17/17 \\
shop & L & L & multi & 8/11 & 20 & 0.0 & 15.0 & 5.0 & \textbf{95.0} & 99.2 & 26 & 19/20\,/\,-- \\
soap to dish & M & S & multi & 3/3 & 20 & 95.0 & 90.0 & \textbf{100.0} & 90.0 & 90.0 & 21 & 5/5\,/\,13/15 \\
teacup to saucer teapot to tray & M & S & multi & 3/3 & 20 & 85.0 & 30.0 & 55.0 & \textbf{95.0} & 97.5 & 7 & 5/5\,/\,14/15 \\
tighten nut & H & L & single & 4/15 & 20 & 10.0 & 15.0 & 10.0 & \textbf{70.0} & 85.0 & 70 & -- \\
utensils to holder & L & S & multi & 5/15 & 20 & 10.0 & 15.0 & 15.0 & \textbf{60.0} & 77.5 & 8 & 4/10\,/\,8/10 \\
\midrule
All tasks &  &  &  & 101/192 & 510 & 46.7 & 53.0 & 56.0 & \textbf{81.7} & 91.2 & 34 &  \\
\bottomrule
\end{tabular}
}
\end{table*}

\begin{table*}[!htbp]
  \centering
  \footnotesize
  \setlength{\tabcolsep}{4pt}
  \caption{SR (\%) and score of \astra with \method on EBench by task group: the first episode of each of the 26 adaptation chains, which reads only the task memory, all 192 training episodes, and the 510 test episodes of Table~\ref{tab:ebench} with the fixed working memory. SR and score are averaged within each task and then across tasks.}
  \label{tab:supp:ebench_train}
  \fitwidth{\begin{tabular}{@{}lrccccccccc@{}}
\toprule
 & & \multicolumn{2}{c}{Overall} & \multicolumn{3}{c}{Precision} & \multicolumn{2}{c}{Mobility} & \multicolumn{2}{c}{Horizon} \\
\cmidrule(lr){3-4}\cmidrule(lr){5-7}\cmidrule(lr){8-9}\cmidrule(l){10-11}
Episodes & $n$ & SR & Score & Low & Med. & High & Mobile & Tabletop & Short & Long \\
\midrule
First training episode (task memory only) & 26 & 57.7 & 72.2 & 85.7 & 37.5 & 0.0 & 73.7 & 14.3 & 72.7 & 46.7 \\
All training episodes (online) & 192 & 70.8 & 83.6 & 88.1 & 58.3 & 35.2 & 80.1 & 45.3 & 81.5 & 62.9 \\
\rowcolor{astralight} Test episodes (fixed working memory) & 510 & 81.7 & 91.2 & 90.1 & 66.7 & 82.5 & 83.2 & 77.9 & 89.1 & 76.3 \\
\bottomrule
\end{tabular}
}
\end{table*}

The seen/unseen partition groups existing test episodes by layout overlap with completed adaptation episodes. Tables~\ref{tab:supp:ebench_pertask} and~\ref{tab:supp:ebench_train} break down the EBench results of Section~\ref{sec:results:ebench} by task and over the adaptation chains, and Table~\ref{tab:generalization} in Section~\ref{sec:results:why} compares seen and unseen layouts. Across the 26 tasks, the fixed working memory improves or matches the SR of direct control on 25 tasks and improves or matches the stronger of the two harnesses with policy on 21 tasks.

\emph{Coordinate replay}, a moving waypoint of Appendix~\ref{sec:supp:replay} within 1\,cm of a coordinate in the fixed working memory, concentrates on the seven single-layout tasks. Over their 140 test episodes, a mean of 66.1\% of the moving waypoints lies within 1\,cm of a stored coordinate, against 19.2\% over the 330 episodes of the 17 multi-layout tasks whose working memory stores coordinates. Scored against the same working memories, the direct-control episodes on the same test scenes reach 4.3\% and 1.0\%, which gives the share expected without working memory. On the other two multi-layout tasks, apple from shelf and dish, the fixed working memory stores no coordinates and reaches 85.0\% and 90.0\% SR.

\paragraph{Adaptation chains.} Table~\ref{tab:supp:ebench_train} shows that in every task group, the SR rises from the first training episode to all training episodes and again to the test episodes. \astra succeeds in the first chain episode on 15 of the 26 tasks but on no high-precision task. The rise from training to test is largest on the high-precision tasks, where the chains run 9 to 15 episodes and the reflection turns mostly analyze failed attempts.

\section{Per-Task Results on RoboDojo}
\label{sec:supp:robodojo}
\begin{table*}[!htbp]
  \centering
  \footnotesize
  \setlength{\tabcolsep}{4pt}
  \caption{Per-task SR (\%) of \astra on the 42 tasks of RoboDojo, grouped by capability dimension. The official zero-shot entry covers 50 layouts per task. Task memory only and \method online run on the same 15 training layouts per task, and the fixed working memory is evaluated on the 50 test layouts per task. On Generalization tasks, each value averages the standard and randomized configurations. The overall SR averages the five dimensions, and bold marks the best value in each row.}
  \label{tab:supp:robodojo_pertask}
  \fitwidth{\begin{tabular}{@{}lrrrr@{}}
\toprule
 & \multicolumn{1}{c}{Official} & \multicolumn{2}{c}{Online, 15 training} & \multicolumn{1}{c}{Fixed} \\
\cmidrule(lr){3-4}\cmidrule(l){5-5}
Task & zero-shot & Task mem. & \method & 50 test \\
\midrule
\rowcolor{tabband} \multicolumn{5}{c}{\textit{Generalization}} \\
\midrule
arrange largest number & 60.0 & 79.5 & 67.0 & \textbf{92.0} \\
fold clothes & 72.0 & 72.3 & 80.4 & \textbf{90.0} \\
hang mugs & 0.0 & 0.0 & \textbf{6.2} & 2.0 \\
make toast & 0.0 & 0.0 & \textbf{20.5} & 16.0 \\
pack objects into box & 0.0 & 0.0 & 7.1 & \textbf{12.0} \\
pour liquid into cup & 4.0 & 12.5 & 46.4 & \textbf{70.0} \\
push T & \textbf{60.0} & 26.8 & 21.4 & 34.0 \\
sort nesting dolls by size & 18.0 & 12.5 & 32.1 & \textbf{60.0} \\
stack blocks & 86.0 & 79.5 & \textbf{100.0} & 98.0 \\
stack bowls & 62.0 & 59.8 & 74.1 & \textbf{88.0} \\
store laptop and headphones & \textbf{0.0} & \textbf{0.0} & \textbf{0.0} & \textbf{0.0} \\
sweep blocks & 4.0 & 6.2 & 46.4 & \textbf{62.0} \\
\midrule
\rowcolor{tabband} \multicolumn{5}{c}{\textit{Precision}} \\
\midrule
build tower & 2.0 & 53.3 & 73.3 & \textbf{94.0} \\
deposit coin & 2.0 & 26.7 & 73.3 & \textbf{92.0} \\
fasten screws & 16.0 & \textbf{33.3} & 13.3 & 28.0 \\
insert key & 0.0 & 0.0 & 6.7 & \textbf{16.0} \\
insert tubes & 0.0 & 13.3 & 40.0 & \textbf{60.0} \\
play xylophone & 8.0 & 66.7 & \textbf{100.0} & 98.0 \\
plug in charger & 0.0 & 0.0 & 40.0 & \textbf{80.0} \\
pour balls into vase & 4.0 & 0.0 & 20.0 & \textbf{48.0} \\
\midrule
\rowcolor{tabband} \multicolumn{5}{c}{\textit{Long-Horizon}} \\
\midrule
classify objects & 56.0 & 60.0 & 86.7 & \textbf{96.0} \\
fill egg holder & 0.0 & 0.0 & 6.7 & \textbf{22.0} \\
fill pen holder & 0.0 & 0.0 & 6.7 & \textbf{14.0} \\
make kong & 0.0 & 33.3 & 86.7 & \textbf{96.0} \\
organize table & 0.0 & 13.3 & 13.3 & \textbf{28.0} \\
play stacking toy & 0.0 & 0.0 & 6.7 & \textbf{24.0} \\
play tic tac toe & 0.0 & 93.3 & \textbf{100.0} & \textbf{100.0} \\
put bottles into dustbin & 10.0 & 26.7 & 80.0 & \textbf{88.0} \\
\midrule
\rowcolor{tabband} \multicolumn{5}{c}{\textit{Memory}} \\
\midrule
cover blocks & 46.0 & 73.3 & \textbf{100.0} & \textbf{100.0} \\
imitate sorting sequence & 36.0 & 40.0 & 73.3 & \textbf{94.0} \\
match and pick from conveyor & 62.0 & 66.7 & 73.3 & \textbf{88.0} \\
press by number & 70.0 & \textbf{100.0} & \textbf{100.0} & \textbf{100.0} \\
swap blocks & 0.0 & 80.0 & \textbf{100.0} & \textbf{100.0} \\
swap T & 18.0 & 46.7 & 73.3 & \textbf{92.0} \\
\midrule
\rowcolor{tabband} \multicolumn{5}{c}{\textit{Open}} \\
\midrule
align blocks & 50.0 & 60.0 & \textbf{100.0} & \textbf{100.0} \\
classify objects by language & 30.0 & 33.3 & \textbf{86.7} & 76.0 \\
general pickup & \textbf{84.0} & 66.7 & 73.3 & \textbf{84.0} \\
pick from conveyor by image & 4.0 & 13.3 & 60.0 & \textbf{72.0} \\
pour by language & 0.0 & 0.0 & 73.3 & \textbf{98.0} \\
solve equation & 40.0 & 40.0 & 80.0 & \textbf{92.0} \\
stack blocks by language & 40.0 & 46.7 & 86.7 & \textbf{98.0} \\
store tools in toolbox & \textbf{0.0} & \textbf{0.0} & \textbf{0.0} & \textbf{0.0} \\
\midrule
Overall (mean of dimensions) & 22.48 & 36.37 & 58.53 & \textbf{69.63} \\
\bottomrule
\end{tabular}
}
\end{table*}

Table~\ref{tab:supp:robodojo_pertask} breaks down Tables~\ref{tab:robodojo} and~\ref{tab:robodojo_stages} by task, and the remainder of this section compares \method with other agentic systems. On the test layouts, the fixed working memory improves or matches the official zero-shot entry of \astra on 41 of the 42 tasks and reaches at least 90\% SR on 18 tasks. On 9 of the 12 Generalization tasks, randomization costs \method at most four SR points or raises its SR. On the 15 training layouts, online adaptation improves or matches the SR of task memory alone on 39 of the 42 tasks.

\subsection{Comparison with Other Agentic Systems}
\label{sec:supp:agentic}
\begin{table}[!htbp]
  \centering
  \footnotesize
  \setlength{\tabcolsep}{2pt}
  \caption{Score (0--100) on the seven RoboDojo tasks evaluated by RAX~\citep{huang2026rax}, with the first three tasks in their randomized configurations. Italic rows come from the RAX paper with at most 16 or 64 exploration episodes per task, and its programs run on metric depth and camera poses outside our information protocol. \method uses the working memory fixed after 15 training episodes and is evaluated on its test layouts.}
  \label{tab:rax}
  \begin{tabular*}{\linewidth}{@{\extracolsep{\fill}}lcccccccc@{}}
\toprule
\hd{System} & \hd{Toast} & \hd{Mugs} & \hd{Sweep} & \hd{Plug} & \hd{Swap} & \hd{Kong} & \hd{Pour} & \hd{Mean} \\
\midrule
\textit{RAX, $\le$16} & \textit{2.5} & \textit{36.9} & \textit{53.3} & \textit{94.0} & \textit{100.0} & \textit{64.0} & \textit{100.0} & \textit{64.4} \\
\textit{RAX, $\le$64} & \textit{47.2} & \textit{39.2} & \textit{71.7} & \textit{94.0} & \textit{100.0} & \textit{96.0} & \textit{100.0} & \textit{78.3} \\
\rowcolor{astralight} \method, 15 & 26.0 & 24.0 & 56.0 & 80.0 & 100.0 & 96.0 & 98.0 & 68.6 \\
\bottomrule
\end{tabular*}

\end{table}

Three agentic systems report RoboDojo results outside the public leaderboard. COAP~\citep{hu2026embodied} executes synthesized code policies on camera parameters and object sizes outside our information protocol and reports 70.24\% SR. RAX~\citep{huang2026rax}, whose coding agent explores each task and submits a program that runs without the agent on metric depth and camera poses, evaluates seven of the 42 tasks and reports scores, which Table~\ref{tab:rax} compares with \method. With at most 16 exploration episodes per task, RAX scores below \method, and with up to 64 it scores higher, mainly on the three randomized tasks and the plug task. SenseNova~\citep{sensenova2026astra} provides \astra with a knowledge package fixed before testing and reports an unweighted task-averaged SR of 27.62\% on five layouts per task under the official step limit at 12.47M tokens per episode. Under the same unweighted average over the 42 tasks on the test layouts, the fixed working memory of \method reaches 66.71\% SR at 1.49M tokens per episode.

\IfFileExists{arxiv/sec/supp/s5_ablations.tex}{\section{Extended Ablations}
\label{sec:supp:ablations}
This section details the protocols and per-task results of the ablations in Section~\ref{sec:analysis}. Appendix~\ref{sec:supp:design} covers memory designs in both agent harnesses, Appendix~\ref{sec:supp:outcome} outcome feedback, Appendix~\ref{sec:supp:trajectory} the executed trajectory of $\pi_{0.5}$, Appendix~\ref{sec:supp:rsi} the comparison with the critic-driven RSI loop, Appendix~\ref{sec:supp:matched} matched controls across layouts, Appendix~\ref{sec:supp:replay} the distance threshold of coordinate replay, and Appendix~\ref{sec:supp:failures} how the working memory fails.

\subsection{Memory Designs in Both Agent Harnesses}
\label{sec:supp:design}
\paragraph{Protocol.} Each design runs one chain of ten validation scenes per task in the same scene order, and the first episode of each chain reads the task memory alone. The designs differ only in the information retained across episodes. The first design reads only the task memory in every episode. The second design carries the text working memory of \method. The third design adds visual replay to the text working memory. The replay archives the frames and robot states of earlier attempts for inspection during execution and reflection. Both agent harnesses run these three designs on the four high-precision tasks. In the harness with $\pi_{0.5}$, a fourth design, described in Appendix~\ref{sec:supp:trajectory}, adds the executed trajectory of every policy call to the text design on the four high-precision tasks. Under direct control, a continuous design adapted from VISTA~\citep{han2026vista} replaces the working memory with one conversation across the ten episodes of each chain, together with the task memory, a guide that the MLLM revises, and an archive of observed frames. Time counts the control phase of an episode, from the first observation until the MLLM ends control, as model time plus 0.5\,s per physics step.

\paragraph{Results.} Tables~\ref{tab:design} and~\ref{tab:harness} list the results. In the harness with $\pi_{0.5}$, the clearest effect of memory concerns completion judgment on collecting coffee beans. With task memory only, the MLLM ends control before the evaluator reports a result in all ten episodes of this task, and with the text working memory and with the visual replay in five of ten. The successes of the task stay at 0 and 1, and its mean score rises from 0 to 31.4 and 27.9. Under direct control, on gear installation, peg insertion, and nut tightening, the agent without working memory already grasps near the poses that the final working memory records, so the working memory mainly changes the actions after the grasp. The MLLM rarely consults the visual replay in either agent harness.

\paragraph{Memory in the harness with $\pi_{0.5}$.} Working memory changes how the MLLM uses the harness with $\pi_{0.5}$ but leaves its success nearly unchanged. Each policy call returns the executed step count, the robot state, and the camera images after the call, and the MLLM keeps the language instruction of the policy unchanged, so the trajectory of the policy stays outside the context of the MLLM. The reflection therefore records when to call the policy and for how many steps, and the same recipe can succeed in one scene and fail in the next. With working memory, the MLLM moves control from the policy to its own end-effector commands. The share of the commanded physics steps that $\pi_{0.5}$ executes falls from 53\% over the first three episodes of each chain to 36\% over the last four, while it stays between 55\% and 60\% with task memory only. These end-effector commands return to coordinates of the working memory about half as often as under direct control. The critic-driven RSI loop of Appendix~\ref{sec:supp:rsi} follows the same pattern, as $\pi_{0.5}$ executes 72.7\% of the commanded steps with the empty playbook and 67.5\% after five rounds.

\subsection{Outcome Feedback}
\label{sec:supp:outcome}

\begin{table}[!htbp]
  \centering
  \footnotesize
  \setlength{\tabcolsep}{3.5pt}
  \caption{SR (\%) and mean score of \astra over the 20 episodes of each of eight chains, with the working memory written without and with the official success and score of each episode. Bold marks the better SR and score in each row.}
  \label{tab:supp:outcome}
  \fitwidth{\begin{tabular}{@{}llcccc@{}}
    \toprule
     & & \multicolumn{2}{c}{\hd{Without outcomes}} & \multicolumn{2}{c}{\hd{With outcomes}} \\
    \cmidrule(lr){3-4}\cmidrule(l){5-6}
    \hd{Task} & \hd{Layouts} & \hd{SR} & \hd{Score} & \hd{SR} & \hd{Score} \\
    \midrule
    collect coffee beans & single & 35.0 & 54.6 & \best{50.0} & \best{71.4} \\
    install gear & single & 25.0 & 35.0 & \best{65.0} & \best{65.0} \\
    peg in hole & single & \best{90.0} & \best{95.0} & 65.0 & 80.0 \\
    tighten nut & single & \best{60.0} & \best{80.0} & 30.0 & 65.0 \\
    dishwasher & multi & \best{95.0} & \best{98.3} & 90.0 & \best{98.3} \\
    utensils to holder & multi & 45.0 & 67.5 & \best{95.0} & \best{97.5} \\
    fruit & multi & 70.0 & 92.0 & \best{95.0} & \best{98.0} \\
    dish & multi & \best{100.0} & \best{100.0} & \best{100.0} & \best{100.0} \\
    \midrule
    Mean, high-precision & & \best{52.5} & 66.2 & \best{52.5} & \best{70.4} \\
    Mean, multi-layout & & 77.5 & 89.5 & \best{95.0} & \best{98.5} \\
    Mean, 8 tasks & & 65.0 & 77.8 & \best{73.8} & \best{84.4} \\
    \bottomrule
  \end{tabular}}
\end{table}

\paragraph{Protocol.} To evaluate how withholding the official success and score affects the evolution of $\Mwork_{k}$ from scratch, eight chains run online adaptation across 20 scenes of their task in a fixed order from an empty working memory, once with the official success and score after each episode and once without them, with the same task memory, prompt, tools, step limits, and size budget. The tasks are the four high-precision tasks, all single-layout, and the multi-layout tasks of dishwasher loading, utensil placement, the fruit task, and dish placement. Without outcome feedback, every episode ends with the same message, the record of earlier attempts omits the success, score, and step count of each attempt, and the reflection prompt states that the official result is not shown.

\paragraph{Results.} In Table~\ref{tab:supp:outcome}, withholding outcomes lowers the mean score over the eight chains from 84.4 to 77.8. The direction varies across tasks, as the SR falls on four, rises on three, and ties on one. On the four high-precision tasks, the SR stays at 52.5\% in both conditions, with peg insertion and nut tightening higher and gear installation and coffee-bean collection lower without outcomes. On the four multi-layout tasks, the SR falls from 95.0\% to 77.5\%, mostly through utensil placement and the fruit task. Over the last five positions of each of these chains, which the matched control of Appendix~\ref{sec:supp:matched} also covers, the mean score is 71.2 with task memory only, 90.0 without outcomes, and 98.0 with outcomes. An episode without outcomes takes 3.29M tokens on average, against 2.67M with outcomes.

\subsection{Executed Trajectory of $\pi_{0.5}$}
\label{sec:supp:trajectory}
As defined in Section~\ref{sec:method:setting}, a policy call reports only the state at its end, so the motion of $\pi_{0.5}$ never enters the trajectory $\xi_k$ on which the reflection operates. The $\pi_{0.5}$ trajectory design in Table~\ref{tab:harness} places this motion into $\xi_k$.

\paragraph{Recorded states.} $\pi_{0.5}$ predicts absolute joint and gripper targets and replans every 30 physics steps. At the end of each replanning interval, the harness reads the proprioception of the robot and records the end-effector pose of each arm, as a position and a unit quaternion in the planning frame of that arm, together with the gripper openings, the base pose, and the physics step. These fields and frames match the measured state that every end-effector command returns, so the MLLM can compare a state reached by $\pi_{0.5}$ with its own waypoints and command that state directly. The last entry of each policy call records the state at the end of that call, and because a physical robot measures these quantities at every control step, the record satisfies the information protocol of Section~\ref{sec:method:setting}.

\paragraph{Exposure to the MLLM.} The record reaches the MLLM through three channels. First, the result of each policy call lists one entry per replanning interval next to the existing step count, state, and images. Second, the memory tools that list earlier calls and measured robot states return the same entries from the episode record, so the MLLM can query them during the episode and during reflection. Third, the reflection prompt after the episode includes the entries of every policy call, so the record remains available for reflection in long episodes. All other settings, including the tool descriptions, the size budget of the working memory, the step limits, and the $\pi_{0.5}$ checkpoint, match the text design.

\paragraph{Results.} With the executed trajectory, the harness succeeds in 17.5\% of the episodes, down from 25.0\%, while the use of the harness stays nearly unchanged. Every policy call of this design returns its executed trajectory, yet $\pi_{0.5}$ executes 41\% of the commanded physics steps, against 42\% with the text design, the final working memories store poses about as often, and the MLLM commands positions near the states that $\pi_{0.5}$ reached about as often as with the text design. These results indicate that visibility of the motion of $\pi_{0.5}$ alone changes neither how the MLLM divides control nor what its working memory records.

\subsection{Comparison with the Critic-Driven RSI Loop}
\label{sec:supp:rsi}

\begin{table}[!htbp]
  \centering
  \footnotesize
  \setlength{\tabcolsep}{4pt}
  \caption{Successes out of 5 on the 90 test episodes shared with the critic-driven RSI loop, for the loop with an empty playbook (Start) and after five rounds (Final), and for the fixed working memory of \method under the train--test protocol. Bold marks the best value in each row.}
  \label{tab:supp:rsi}
  \fitwidth{\begin{tabular}{@{}lccc@{}}
    \toprule
    \hd{Task} & \hd{Start} & \hd{Final} & \hd{\method} \\
    \midrule
    apple to fruit bowl & \best{5} & \best{5} & \best{5} \\
    bottle & 0 & 0 & \best{3} \\
    bowl to plate & \best{5} & 4 & \best{5} \\
    collect coffee beans & 0 & 2 & \best{5} \\
    dishwasher & 2 & \best{4} & \best{4} \\
    flip cup collect cookies & \best{2} & \best{2} & \best{2} \\
    install gear & 1 & 2 & \best{4} \\
    make sandwich & 4 & \best{5} & 3 \\
    microwave & 2 & 3 & \best{4} \\
    peg in hole & 0 & 0 & \best{3} \\
    pen & \best{1} & 0 & \best{1} \\
    perfume to cosmetics rack & \best{5} & 4 & \best{5} \\
    put glass in glassbox & 3 & \best{4} & \best{4} \\
    shop & 0 & 0 & \best{4} \\
    soap to dish & \best{5} & \best{5} & \best{5} \\
    teacup to saucer teapot to tray & 3 & 2 & \best{5} \\
    tighten nut & 1 & 0 & \best{3} \\
    utensils to holder & 1 & 0 & \best{4} \\
    \midrule
    Successes of 90 & 40 & 42 & \best{69} \\
    SR (\%) & 44.4 & 46.7 & \best{76.7} \\
    Mean score & 61.3 & 64.4 & \best{92.1} \\
    \bottomrule
  \end{tabular}}
\end{table}

\paragraph{Protocol.} The critic-driven RSI loop starts from the harness with $\pi_{0.5}$ and an empty playbook. In each of five rounds, ten training tasks run once with and once without their demonstration, 100 episodes in all, and a read-only critic instance of the same model inspects these episodes and accepts revisions to the playbook and the harness. Its starting point (Start) and its final version (Final) are evaluated on the first five test scenes of each of 18 tasks. The fixed working memory of the train--test protocol is scored on the same 90 scenes.

\paragraph{Results.} As Table~\ref{tab:supp:rsi} shows, the fixed working memory reaches a mean score of 92.1, against 64.4 for Final and 61.3 for Start, and matches or exceeds Final on 17 of the 18 tasks. The largest differences arise on shop checkout and utensil placement, where Final fails in all five episodes and the fixed working memory succeeds in four.

\subsection{Matched Controls per Task}
\label{sec:supp:matched}

\begin{table}[!htbp]
  \centering
  \footnotesize
  \setlength{\tabcolsep}{4pt}
  \caption{Successes out of 5 on the last five test episodes of each multi-layout EBench task with task memory only and with the fixed working memory of \method, paired by scene. Bold marks the better value in each row.}
  \label{tab:supp:matched_eb}
  \fitwidth{\begin{tabular}{@{}lcc@{}}
    \toprule
    \hd{Task} & \hd{Task memory only} & \hd{\method} \\
    \midrule
    apple from shelf & \best{5} & \best{5} \\
    apple to fruit bowl & \best{5} & \best{5} \\
    bookmark on book & \best{5} & 4 \\
    bottle & 0 & \best{2} \\
    bowl to plate & 4 & \best{5} \\
    detergent & 4 & \best{5} \\
    dish & 4 & \best{5} \\
    dishwasher & 0 & \best{5} \\
    fruit & 2 & \best{5} \\
    make sandwich & 2 & \best{3} \\
    microwave & \best{4} & \best{4} \\
    pen & \best{1} & 0 \\
    perfume to cosmetics rack & \best{5} & \best{5} \\
    remote to holder & \best{5} & \best{5} \\
    salt to spice rack & \best{5} & \best{5} \\
    shop & 2 & \best{4} \\
    soap to dish & \best{5} & \best{5} \\
    teacup to saucer teapot to tray & \best{5} & \best{5} \\
    utensils to holder & 1 & \best{4} \\
    \midrule
    Successes of 95 & 64 & \best{81} \\
    SR (\%) & 67.4 & \best{85.3} \\
    \bottomrule
  \end{tabular}}
\end{table}

\begin{table*}[!htbp]
  \centering
  \scriptsize
  \setlength{\tabcolsep}{2.5pt}
  \caption{Successes out of 3 on the first three RoboDojo test layouts of each task in native order (126 episodes), with task memory only (TM) and with the fixed working memory of \method (WM), paired by layout. Gen., Prec., Long, Mem., and Open denote the five capability dimensions.}
  \label{tab:supp:matched_rd}
  \fitwidth{\begin{tabular}{@{}llcc@{\hspace{10pt}}llcc@{\hspace{10pt}}llcc@{}}
    \toprule
    \hd{Task} & \hd{Dim.} & \hd{TM} & \hd{WM} & \hd{Task} & \hd{Dim.} & \hd{TM} & \hd{WM} & \hd{Task} & \hd{Dim.} & \hd{TM} & \hd{WM} \\
    \midrule
    arrange largest number & Gen. & 0 & 3 & build tower & Prec. & 2 & 2 & cover blocks & Mem. & 3 & 3 \\
    fold clothes & Gen. & 2 & 3 & deposit coin & Prec. & 2 & 2 & imitate sorting sequence & Mem. & 1 & 2 \\
    hang mugs & Gen. & 0 & 0 & fasten screws & Prec. & 1 & 0 & match and pick from conveyor & Mem. & 3 & 3 \\
    make toast & Gen. & 0 & 1 & insert key & Prec. & 0 & 0 & press by number & Mem. & 3 & 3 \\
    pack objects into box & Gen. & 0 & 0 & insert tubes & Prec. & 1 & 1 & swap T & Mem. & 0 & 3 \\
    pour liquid into cup & Gen. & 3 & 2 & play Xylophone & Prec. & 1 & 3 & swap blocks & Mem. & 3 & 3 \\
    push T & Gen. & 1 & 0 & plug in charger & Prec. & 0 & 1 & align blocks & Open & 2 & 3 \\
    sort nesting dolls by size & Gen. & 0 & 3 & pour balls into vase & Prec. & 0 & 2 & classify objects by language & Open & 2 & 2 \\
    stack blocks & Gen. & 3 & 3 & classify objects & Long & 2 & 3 & general pickup & Open & 3 & 2 \\
    stack bowls & Gen. & 1 & 3 & fill egg holder & Long & 0 & 1 & pick from conveyor by image & Open & 0 & 2 \\
    store laptop and headphones & Gen. & 0 & 0 & fill pen holder & Long & 0 & 0 & pour by language & Open & 0 & 3 \\
    sweep blocks & Gen. & 0 & 1 & make kong & Long & 0 & 3 & solve equation & Open & 2 & 2 \\
     & & & & organize table & Long & 0 & 1 & stack blocks by language & Open & 1 & 3 \\
     & & & & play stacking toy & Long & 0 & 2 & store tools in toolbox & Open & 0 & 0 \\
     & & & & play tic tac toe & Long & 3 & 3 & & & & \\
     & & & & put bottles into dustbin & Long & 3 & 3 & & & & \\
    \midrule
    \multicolumn{12}{@{}l}{SR over the five dimensions: 40.83\% with task memory only and 66.11\% with the fixed working memory.} \\
    \bottomrule
  \end{tabular}}
\end{table*}

\paragraph{EBench.} The matched control runs \astra with task memory only on the last five test episodes of each of the 19 multi-layout tasks and pairs each episode with the episode of the fixed working memory on the same scene, with the results in Table~\ref{tab:supp:matched_eb}. The fixed working memory raises the mean score from 84.5 to 94.7. The largest gains occur on dishwasher loading (from 0 to 5), the fruit task (from 2 to 5), and utensil placement (from 1 to 4). It also lowers the mean tokens per episode from 3.82M to 2.32M.

\paragraph{RoboDojo.} This control uses the first three test layouts of each task in native order and pairs task memory only with the fixed working memory on each layout, as Table~\ref{tab:supp:matched_rd} reports. The fixed working memory reaches a higher SR on 20 tasks and a lower one on 4, and 18 tasks tie. Five tasks rise from 0 to 3 of 3: arranging the largest number, sorting nesting dolls, making a kong, swapping a T-shaped block, and pouring by language.

\subsection{Coordinate Replay and Its Threshold}
\label{sec:supp:replay}

\begin{table}[!htbp]
  \centering
  \footnotesize
  \setlength{\tabcolsep}{2.5pt}
  \caption{Median share of moving waypoints per EBench test episode that lie within 1, 2, or 3\,cm of a coordinate in the fixed working memory. Values in parentheses give the baseline floor obtained by scoring the corresponding direct-control episode (which reads no working memory) against the same stored coordinates. Rows report successful episodes on single-layout tasks, successful episodes on seen and unseen layouts of multi-layout tasks, and all 510 test episodes.}
  \label{tab:supp:replay}
  \fitwidth{\begin{tabular}{@{}lcccc@{}}
    \toprule
    \hd{Episodes} & $n$ & \hd{1\,cm} & \hd{2\,cm} & \hd{3\,cm} \\
    \midrule
    Single layout, succ. & 109 & 0.76 (0.03) & 0.82 (0.15) & 0.88 (0.34) \\
    Seen layout, succ. & 93 & 0.22 (0.00) & 0.45 (0.07) & 0.66 (0.16) \\
    Unseen layout, succ. & 215 & 0.11 (0.00) & 0.28 (0.00) & 0.45 (0.05) \\
    All episodes & 510 & 0.25 (0.00) & 0.45 (0.07) & 0.63 (0.18) \\
    \bottomrule
  \end{tabular}}
\end{table}

Table~\ref{tab:supp:replay} counts as moving waypoints the arm targets that shift by more than 1\,cm from the preceding target of the same arm. On single-layout tasks, successful episodes replay a median of 76\% of their moving waypoints within 1\,cm of a stored coordinate, far above the 3\% baseline floor of direct control without working memory, and this gap holds at 2 and 3\,cm. By contrast, on unseen layouts of multi-layout tasks, the median replay share at 1\,cm drops to 11\%.

\subsection{Failure Modes of the Working Memory}
\label{sec:supp:failures}
Most failures with the fixed working memory follow two patterns, a stored recipe that reproduces failing executions as well as successful ones, and a lesson drawn from training successes whose cause the camera views leave ambiguous. Of the 93 failures among the 510 test episodes of the train--test protocol, 41 replay stored coordinates, with at least a quarter of their moving waypoints within 1\,cm of a coordinate in the fixed working memory, and 27 of these 41 failures occur on single-layout tasks. On nut tightening, each of the six failures places more than half of its moving waypoints near a stored coordinate, as the 14 successes do. Pen placement shows the second case. In four of its seven training successes, the official success arrived during a late adjustment of a pen that already appeared to stand in the holder, and the working memory states that ``a late small intervention can precede success without proving its exact mechanism'' while still recommending such adjustments. With this working memory, pen placement succeeds in 5 of 20 test episodes, and each of its 15 failures replays at most 5\% of its moving waypoints. Pen placement and utensil placement together account for 23 of the 52 failures without replay.
}{}
\IfFileExists{arxiv/sec/supp/s6_transfer.tex}{\section{Transfer Protocol}
\label{sec:supp:transfer}

\begin{table}[!htbp]
  \centering
  \footnotesize
  \setlength{\tabcolsep}{2.5pt}
  \caption{SR (\%) and mean score of four MLLMs on two EBench tasks, 20 episodes each, with task memory only and with the working memory that \astra wrote at the same chain position.}
  \label{tab:supp:transfer}
  \begin{tabularx}{\linewidth}{l*{4}{>{\centering\arraybackslash}X}}
    \toprule
    & \multicolumn{2}{c}{\hd{Task memory only}} & \multicolumn{2}{c}{\hd{Task + working memory}} \\
    \cmidrule(lr){2-3}\cmidrule(lr){4-5}
    \hd{Model} & \hd{SR} & \hd{Score} & \hd{SR} & \hd{Score} \\
    \midrule
    \rowcolor{tabband} \multicolumn{5}{c}{\textit{Put glasses into a glassbox}} \\
    \midrule
    Gemini 3.8 Flash & 0.0 & 15.0 & \best{90.0} & \best{91.7} \\
    MiMo-V2.6-Pro & 0.0 & 6.7 & 70.0 & 78.3 \\
    DeepSeek-V4.1-Flash & 0.0 & 6.7 & 40.0 & 45.0 \\
    GLM-5.3-Flash & 0.0 & 3.3 & 20.0 & 25.0 \\
    \midrule
    \rowcolor{tabband} \multicolumn{5}{c}{\textit{Place utensils in a holder}} \\
    \midrule
    Gemini 3.8 Flash & 20.0 & 40.0 & \best{35.0} & \best{57.5} \\
    MiMo-V2.6-Pro & 0.0 & 0.0 & 10.0 & 12.5 \\
    DeepSeek-V4.1-Flash & 0.0 & 10.0 & 5.0 & 20.0 \\
    GLM-5.3-Flash & 0.0 & 5.0 & 5.0 & 10.0 \\
    \bottomrule
  \end{tabularx}
\end{table}

\paragraph{Mixed generalization setting.} For Table~\ref{tab:transfer}, each MLLM attempts the five test episodes in the Mixed generalization setting of every EBench task in two independent direct-control sessions. One session reads the task memory only, and the paired session reads the fixed working memory that \astra reads in Table~\ref{tab:ebench}. All MLLMs receive the same observation space, action tools, task memory, and episode prompt. The gains hold across task layouts. On the 35 single-layout episodes, the working memory lifts Gemini 3.8 Flash from 5 to 18 successes, DeepSeek-V4.1-Flash from 0 to 2, and \astra from 9 to 26. On the 95 multi-layout episodes, it lifts them from 26 to 52, from 1 to 15, and from 70 to 80, respectively.

\paragraph{Chain study with four MLLMs.} To examine how transfer develops as the working memory accumulates experience from scratch, four recipient MLLMs attempt 20 scenes of putting glasses into a glassbox and of placing utensils in a holder twice in independent direct-control sessions. At chain position $k \in \{1,\dots,20\}$, one session reads the task memory $\Mtask$ only, and the paired session additionally reads the working memory snapshot $\Mwork_{k-1}$ that \astra wrote after episode $k-1$ along its 20-episode adaptation chain on that task, so both sessions of the first position read the task memory alone. As Table~\ref{tab:supp:transfer} shows, the working memory improves all four recipients, with Gemini 3.8 Flash reaching 25 of 40 successes, up from 4, and MiMo-V2.6-Pro reaching 16, up from 0.

\paragraph{Replay by recipients.} The replay share of a recipient episode is the fraction of its moving waypoints, defined as in Appendix~\ref{sec:supp:replay}, that lie within 1\,cm of a coordinate written in the working memory of its position. The median share on glasses packing is 0.61 for Gemini 3.8 Flash, 0.59 for MiMo-V2.6-Pro, 0.27 for GLM-5.3-Flash, and 0.17 for DeepSeek-V4.1-Flash, against at most 0.05 for the same models with task memory only, measured against the same coordinates. On utensil placement, no recipient episode reaches a share of 0.5, yet following the distilled lessons still raises the SR of four recipients.

\paragraph{Low-level gripper behavior.} The gripper statistic of Section~\ref{sec:analysis:transfer} pairs DeepSeek-V4.1-Flash with and without the working memory that \astra wrote after 15 episodes of its adaptation chain, on 24 episodes of eight tasks. Among the waypoints that raise the hand, it counts those with a gripper target between closed and fully open, a setting that holds an object at its measured width without squeezing it.
}{}
\IfFileExists{arxiv/sec/supp/s7_cost.tex}{\section{Cost and Wall-Clock Time}
\label{sec:supp:cost}

\begin{table*}[!htbp]
  \centering
  \footnotesize
  \setlength{\tabcolsep}{4pt}
  \caption{Mean token composition per episode of \astra on EBench, with the cached share of input tokens, uncached input and output tokens per episode, and tokens per success, which divide all tokens of an evaluation, failed episodes included, by its successes. The training row covers the 192 episodes of the 26 adaptation chains, reflection turns included.}
  \label{tab:supp:cost_eb}
  \fitwidth{\begin{tabular}{@{}lcccccccc@{}}
    \toprule
    \hd{Configuration} & \hd{Episodes} & \hd{Successes} & \shortstack{\hd{Tokens/ep.}\\\hd{mean (M)}} & \shortstack{\hd{Tokens/ep.}\\\hd{median (M)}} & \shortstack{\hd{Cached}\\\hd{input (\%)}} & \shortstack{\hd{Uncached}\\\hd{input (k)}} & \shortstack{\hd{Output}\\\hd{(k)}} & \shortstack{\hd{Tokens/}\\\hd{success (M)}} \\
    \midrule
    Direct control & 510 & 237 & 3.77 & 3.12 & 91.5 & 75.6 & 11.2 & 8.12 \\
    Harness with $\pi_{0.5}$ & 510 & 269 & 1.83 & 1.34 & 96.5 & 64.2 & 4.6 & 3.46 \\
    Harness with InternW0-$\Delta$ & 510 & 284 & 2.22 & 1.76 & 97.0 & 65.7 & 4.9 & 3.99 \\
    \method, fixed working memory (test) & 510 & 417 & 2.26 & 1.38 & 97.1 & 64.3 & 7.6 & 2.76 \\
    \method, adaptation chains (training) & 192 & 101 & 4.09 & 3.47 & 97.1 & 118.1 & 14.4 & 7.77 \\
    \bottomrule
  \end{tabular}}
\end{table*}

\begin{table}[!htbp]
  \centering
  \footnotesize
  \setlength{\tabcolsep}{2.5pt}
  \caption{Median wall-clock minutes per EBench test episode by task group, measured from the start to the end of agent control, excluding simulator queueing and scene resets. Bold marks the shortest median in each group.}
  \label{tab:supp:wall}
  \fitwidth{\begin{tabular}{@{}lccccc@{}}
    \toprule
    \hd{Group} & $n$ & \hd{Direct} & \hd{$\pi_{0.5}$} & \hd{InternW0-$\Delta$} & \hd{\method} \\
    \midrule
    All & 510 & 36.7 & 23.9 & 36.5 & \best{14.3} \\
    Tabletop & 140 & 52.5 & 27.5 & 50.0 & \best{15.1} \\
    Mobile & 370 & 29.0 & 21.3 & 32.0 & \best{13.8} \\
    Short horizon & 220 & 20.6 & 12.7 & 20.5 & \best{7.4} \\
    Long horizon & 290 & 53.5 & 33.3 & 63.2 & \best{19.9} \\
    High precision & 80 & 48.2 & 24.7 & 35.4 & \best{13.8} \\
    \bottomrule
  \end{tabular}}
\end{table}

\paragraph{Tokens on EBench.} As Table~\ref{tab:supp:cost_eb} shows, prompt caching dominates the token composition of every configuration. Because the agent resends its growing context at every turn of an episode, 91.5\% to 97.1\% of the input tokens repeat an earlier prefix and are served from the cache. The fixed working memory spends 2.26M tokens per test episode, more than the two harnesses with policy, in which one policy call advances many physics steps without an agent turn, and fewer than direct control.

\paragraph{Tokens on RoboDojo.} Table~\ref{tab:supp:cost_rd} gives the corresponding values on RoboDojo. The fixed working memory needs no reflection turn, which keeps its tokens per episode below those of online adaptation, 1.49M against 1.73M.

\begin{table}[!htbp]
  \centering
  \footnotesize
  \setlength{\tabcolsep}{2.5pt}
  \caption{Tokens of \astra on RoboDojo per episode and per success, for task memory only and online adaptation on the 630 episodes of the 15 training layouts and for the fixed working memory on the 2{,}100 test episodes, with the cached share of input tokens and the share of reflection turns. For task memory only, the tokens cover the control phase of each episode. Tokens per success follow Table~\ref{tab:supp:cost_eb}.}
  \label{tab:supp:cost_rd}
  \fitwidth{\begin{tabular}{@{}lcccccc@{}}
    \toprule
    \hd{Configuration} & \hd{Eps.} & \hd{Succ.} & \shortstack{\hd{Tokens/}\\\hd{ep.\ (M)}} & \shortstack{\hd{Cached}\\\hd{(\%)}} & \shortstack{\hd{Reflec-}\\\hd{tion (\%)}} & \shortstack{\hd{Tokens/}\\\hd{succ.\ (M)}} \\
    \midrule
    Task memory only & 630 & 216 & 1.41 & 95.3 & -- & 4.12 \\
    Online adaptation & 630 & 350 & 1.73 & 95.7 & 11.9 & 3.11 \\
    Fixed working memory (test) & 2{,}100 & 1{,}401 & 1.49 & 95.7 & -- & 2.23 \\
    \bottomrule
  \end{tabular}}
\end{table}

\paragraph{Wall-clock time.} On EBench, the fixed working memory has the shortest median episode duration in every task group, 14.3 minutes over all test episodes, compared with 23.9 minutes for the harness with $\pi_{0.5}$ and 36.7 minutes under direct control, as listed in Table~\ref{tab:supp:wall}. On the 2{,}100 RoboDojo test episodes, the median episode with the fixed working memory takes 7.0 minutes, most of it in model turns.
}{}
\IfFileExists{arxiv/sec/supp/s8_cases.tex}{\section{Case Studies}
\label{sec:supp:cases}
This section extends Figure~\ref{fig:cases} of Section~\ref{sec:results:why} and the failure analysis of Appendix~\ref{sec:supp:failures} with further test episodes of \method with the fixed working memory from the train--test evaluations of Tables~\ref{tab:ebench} and~\ref{tab:robodojo}. Figure~\ref{fig:supp:cases_replay} shows coordinate replay on single-layout tasks, Figure~\ref{fig:supp:cases_rules} lesson-guided execution on unseen layouts, and Figure~\ref{fig:supp:cases_failures} three failure modes. Every memory excerpt and agent message in the figures is copied verbatim from the episode record.

\begin{figure*}[!htbp]
  \centering
  \includegraphics[width=\linewidth]{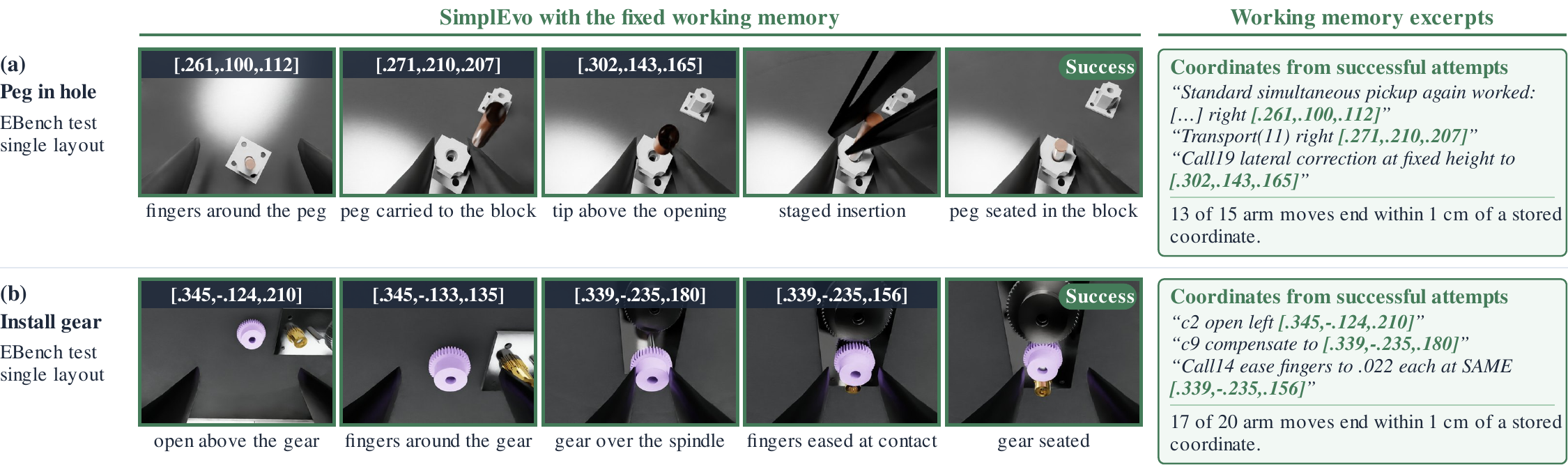}
  \caption{Coordinate replay on two high-precision single-layout tasks of EBench, peg insertion and gear installation. Both episodes of \method with the fixed working memory (green) succeed, and direct control and the harness with $\pi_{0.5}$ fail on the same scenes. Dark labels indicate commanded positions in meters that match coordinates in the working memory, the cards quote the lines that store them, and [\ldots] marks internal omissions. Following Appendix~\ref{sec:supp:replay}, arm moves count the moving commanded waypoints of the episode, whose target lies more than 1\,cm from the previous target of the same arm. Frames come from the wrist cameras, and the last frame shows the terminal state.}
  \label{fig:supp:cases_replay}
\end{figure*}

\paragraph{Coordinate replay.} On single-layout tasks, the coordinates stored from successful training attempts carry a test episode from the grasp to the final insertion, as Figure~\ref{fig:supp:cases_replay} shows. In the peg-insertion episode, 13 of the 15 arm moves end within 1\,cm of a stored coordinate, covering the peg grasp, the transport to the held block, and the lateral correction above the hole. The gear-installation episode reissues the stored poses for opening the gripper above the gear, aligning the gear over the spindle, and easing the fingers at contact, with 17 of its 20 arm moves matching a stored coordinate within 1\,cm.

\begin{figure*}[!htbp]
  \centering
  \includegraphics[width=\linewidth]{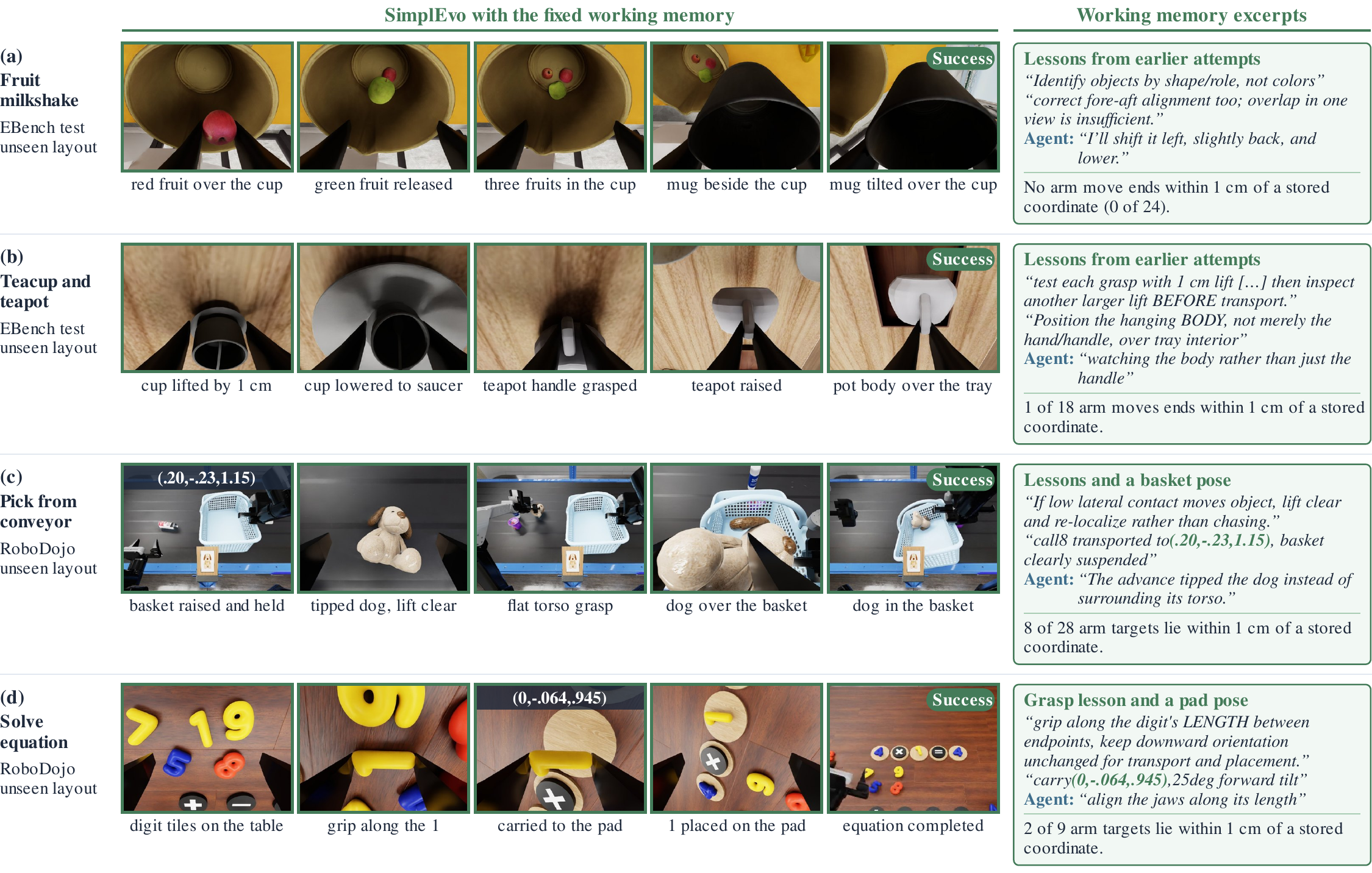}
  \caption{Lessons on unseen layouts. (a, b)~EBench test scenes of the fruit milkshake task and of the teacup and teapot task, whose layouts appear in no training episode of the chain. (c, d)~RoboDojo test layouts of picking from a conveyor by image and of solving an equation. All four episodes of \method with the fixed working memory (green) succeed. On the same scenes, direct control and the harness with $\pi_{0.5}$ fail in (a) and (b), and \astra with task memory only fails in (c) and (d). The cards quote lessons of the working memory read in the episode, and \emph{Agent} marks a message of the agent in the same episode. Dark labels give commanded positions in meters that equal stored coordinates. EBench counts cover the moving commanded waypoints, and RoboDojo counts cover the arm targets away from the home pose. Frames come from the wrist cameras, and in (c) and (d) also from the head camera.}
  \label{fig:supp:cases_rules}
\end{figure*}

\paragraph{Lessons on unseen layouts.} When the test layout differs from every training scene, most commanded positions lie farther than 1\,cm from every stored coordinate, and the episodes follow the lessons of the working memory, as Figure~\ref{fig:supp:cases_rules} illustrates. On the unseen fruit-milkshake layout, none of the 24 arm moves lies within 1\,cm of a stored pose. After the first tilt leaves the rim of the mug right of the cup opening, the agent shifts the mug left, slightly back, and lower, as the stored lesson on fore-aft alignment advises, and the episode succeeds. On the unseen teacup layout, the agent lifts the cup by 1\,cm before transport and lowers the teapot while monitoring the pot body, following two stored lessons. On the unseen conveyor layout, an initial approach tips the dog, and the agent retracts and locates the dog again before a flat torso grasp, as a stored recovery lesson advises. On the unseen equation layout, the agent grips the digit tile along its length, following the stored grasp lesson, and carries it at the stored pad pose.

\begin{figure*}[!htbp]
  \centering
  \includegraphics[width=\linewidth]{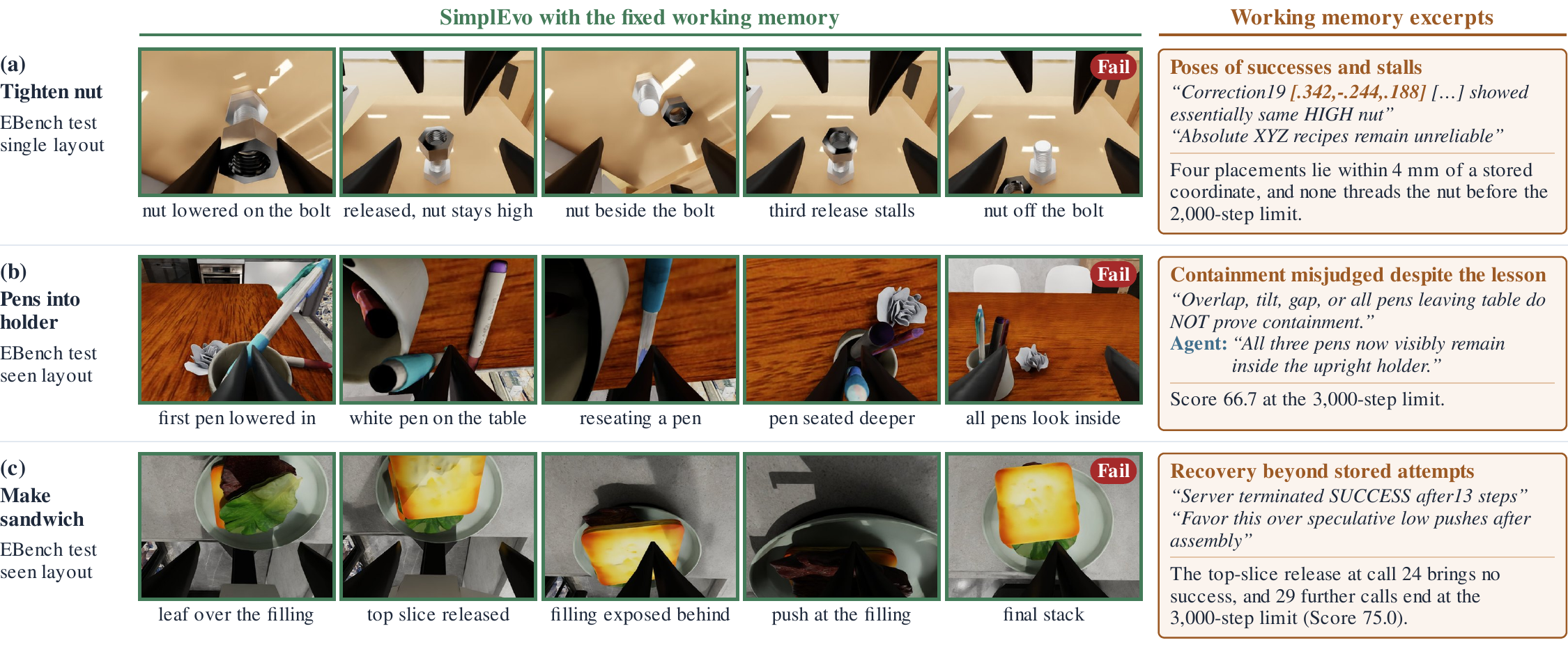}
  \caption{Failures of \method with the fixed working memory on three EBench tasks: nut tightening (score 50.0), pens into a holder (score 66.7), and sandwich making (score 75.0). Each episode ends at the physics step limit of its task. The cards quote the working memory read in the episode, and \emph{Agent} marks the last message of the agent in the episode. Frames come from the wrist cameras.}
  \label{fig:supp:cases_failures}
\end{figure*}

\paragraph{Failure modes.} The three failures in Figure~\ref{fig:supp:cases_failures} arise from stored coordinates that do not separate successful from stalled attempts, from a visual completion judgment that contradicts a stored lesson, and from a post-assembly state that needs a recovery absent from the stored attempts. On nut tightening, the working memory stores seating poses of successful and stalled training attempts a few millimeters apart and also notes that absolute XYZ recipes remain unreliable. The test episode places the nut repeatedly within a few millimeters of a stored coordinate, and no placement threads it before the step limit. On pen placement, the working memory states that overlap, tilt, or a gap do not prove containment, yet the last message of the agent reports all three pens inside the holder, and the episode ends at the step limit with a score of 66.7. On sandwich making, the three latest successful training attempts in the working memory end when the top slice is released. In the test episode, the release of the top slice does not complete the task, and the agent spends the rest of the episode on inspection and corrective pushes, against the stored advice to avoid speculative low pushes after assembly.
}{}

\FloatBarrier

\end{document}